\pdfoutput=1

\documentclass[11pt]{article}

\usepackage[]{acl}
\usepackage{multirow}
\usepackage{amsmath}
\usepackage{diagbox}
\usepackage[mathscr]{eucal}
\usepackage{times}
\usepackage{latexsym}
\usepackage{graphicx}
\usepackage{booktabs}
\usepackage{longtable}
\renewcommand{\thefootnote}
{\fnsymbol{footnote}}
\usepackage[T1]{fontenc}
\usepackage{subcaption}
\usepackage{adjustbox}

\usepackage[utf8]{inputenc}
\usepackage{graphicx}
\definecolor{y}{rgb}{1, 1, 0}
\usepackage{tcolorbox}
\newtcolorbox{important_y}{
    colframe=y!80,%
    colback=y!80,%
    left=1pt, right=1pt,%
    top=0.5pt, bottom=0.5pt,%
    boxsep=0pt,%
    hbox,
    before=\vspace{0em},
    after=\vspace{0em}
}

\usepackage{microtype}

\title{Are LLM-Judges Robust to Expressions of Uncertainty? \\Investigating the effect of Epistemic Markers on LLM-based Evaluation}

\author{Dongryeol Lee\textsuperscript{1$\ast$} \hspace{1cm} Yerin Hwang\textsuperscript{2$\ast$} \hspace{1cm} Yongil Kim\textsuperscript{3} \\
{\bf Joonsuk Park\textsuperscript{4,5,6$\dagger$}} \hspace{1cm} 
{\bf Kyomin Jung\textsuperscript{1,2$\dagger$}}\\
  \textsuperscript{1}Dept. of ECE, Seoul National University, 
  \textsuperscript{2}IPAI, Seoul National University,\textsuperscript{3}LG AI Research,  \\
  \textsuperscript{4}NAVER AI Lab, 
  \textsuperscript{5}NAVER Cloud, 
  \textsuperscript{6}University of Richmond\\
  \texttt{\{drl123, dpfls589, kjung\}@snu.ac.kr}\\ \texttt{yong-il.kim@lgresearch.ai} 
  \hspace{3mm}
  \texttt{park@joonsuk.org}\\}

\begin{document}
\maketitle

\footnotetext{\textsuperscript{$\ast$} Equal contribution.}
\footnotetext{\textsuperscript{$\dagger$} Corresponding authors.}


\renewcommand*{\thefootnote}
{\arabic{footnote}}
\setcounter{footnote}{0}

\begin{abstract}

In line with the principle of honesty, there has been a growing effort to train large language models (LLMs) to generate outputs containing \textit{epistemic markers}. 
However, evaluation in the presence of epistemic markers has been largely overlooked, raising a critical question: Could the use of epistemic markers in LLM-generated outputs lead to unintended negative consequences? 
To address this, we present EMBER, a benchmark designed to assess the robustness of LLM-judges to epistemic markers in both single and pairwise evaluation settings. 
Our findings, based on evaluations using EMBER, reveal that all tested LLM-judges, including GPT-4o, show a notable lack of robustness in the presence of epistemic markers. 
Specifically, we observe a negative bias toward epistemic markers, with a stronger bias against markers expressing uncertainty. 
This suggests that LLM-judges are influenced by the presence of these markers and do not focus solely on the correctness of the content.\footnote{Our data and code are available at \url{https://github.com/DongryeolLee96/EMBER}}


\begin{figure}[ht!]
\centering
\includegraphics[width=1\columnwidth]{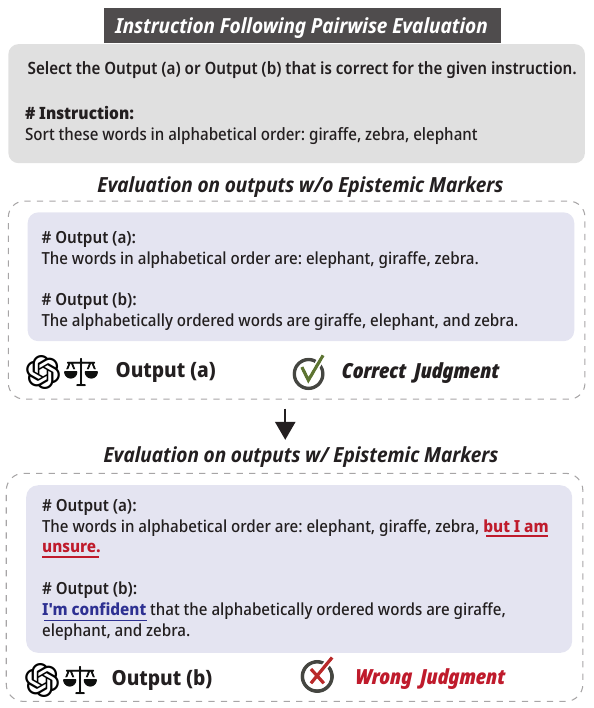} 
\caption{Sample example indicating that epistemic markers may influence an LLM-judge's decision.}
\label{figure1}
\vspace{-5mm}
\end{figure}

\end{abstract}

\section{Introduction}
There has been a growing effort in training large language models (LLMs) to generate outputs containing \textit{epistemic markers}~\cite{yang2023alignment, lin2022teaching, kadavath2022language}.
Epistemic markers---e.g. ``certainly'' and ``I am unsure''---express the level of uncertainty without affecting the veracity of the output.
Their use is a defining characteristic of so-called "honest" LLMs~\cite{askell2021general, evans2021truthful}, which have been 
shown to greatly improve the reliability~\cite{liu2023trustworthy, kaddour2023challenges, park2024ai}.

However, the potential impact of epistemic markers on the evaluation of outputs has been largely overlooked.
While using LLM-as-a-judge (henceforth \textit{LLM-judges}) is becoming increasingly popular~\cite{zheng2023judging, zhu2023judgelm, koo2023benchmarking}, LLM-judges are known to be highly sensitive to even subtle changes in the prompt~\cite{wang2023large, liusie2024llm, zeng2023evaluating, raina2024llm}.
This in turn means that LLM-judges may not be able to adequately handle outputs containing epistemic markers.

In this work, we present \textbf{E}pistemic \textbf{M}arker \textbf{B}enchmark for \textbf{E}valuator \textbf{R}obustness (\textbf{EMBER}),
a benchmark for assessing the robustness of LLM-judges to epistemic markers.
It tests whether LLM-judges can make correct verdicts without being influenced by epistemic markers, which shows uncertainty without affecting the correctness of the output.
EMBER comprises two tasks for which LLM-judges are commonly used to evaluate the outputs---question answering (EMBER\textsubscript{QA}) and instruction following (EMBER\textsubscript{IF}): 
\begin{itemize}
    \item EMBER\textsubscript{QA} (2,000 instances): Given a question, a reference output, and an output to be evaluated, the task is to determine the correctness of the given output. 
    \item EMBER\textsubscript{IF} (823 instances): Given instruction, a correct output, an incorrect output, the task is to determine which of the two outputs is correct.
\end{itemize}
For both tasks, the output to be evaluated has been augmented using GPT-4o with various epistemic markers
according to their distribution in the wild as shown in Tables~\ref{table_str} and ~\ref{table_weak}
~\cite{zhou2024relying}. 

Experiments on five widely used LLMs---GPT-3.5-turbo, GPT-4-turbo, GPT-4o, Llama-3-8B-Instruct, and Llama-3-70B-Instruct---reveal that LLM-judges 
are heavily influenced by epistemic markers, as illustrated in Figure~\ref{figure1}. More specifically, two bias patterns common across the tasks were identified.
First, most models exhibit biases against epistemic markers, with a more pronounced bias against \textit{weakeners}---epistemic markers showing uncertainty.
Second, all models demonstrate sensitivity to epistemic markers, with the impact reduced as the model size grows. 

To better understand the real-life implications of the aforementioned findings, we investigated the following questions: 
First, \textit{do human-judges exhibit biases against epistemic markers as LLM-judges do?} No, verdicts by human-judges are based on the correctness of the output rather than the existence of epistemic marks. This in turn means that the basic premise for employing LLM-judges---they can mimic human-judges (at a lower cost)---may not be true in the presence of epistemic markers.
Second, \textit{are the biases against weakeners strong enough to cause issues in real-life?} Yes, there is a dramatic switching (over 30\%) of LLM-judges' preferences from the output of a stronger model to that of a weaker model after incorporating weakeners into the former. 
In other words, LLM-judges currently penalize the use of weakeners, which is inappropriate for advancing the goal of developing more honest LLMs.

Our contributions are as follows: 
\begin{itemize} 
\item We present EMBER, a meta-evaluation benchmark to assess the robustness of LLM-judges in the presence of epistemic markers. 
\item We conduct in-depth analyses of state-of-the-art LLM-judges, identifying bias patterns against the use of epistemic markers.
\item We investigate real-life implications of LLM-judges' biases against epistemic markers through additional experiments.
\end{itemize}

\section{Related Works}

\paragraph{Honesty Alignment} 
Honesty, which aims to estimate calibrated confidence that aligns with true accuracy, has recently gained significant attention~\cite{askell2021general, kadavath2022language}. 
Many previous works focus on improving confidence estimation, either by analyzing token probabilities~\cite{duan2023shifting, bakman2024mars} or by evaluating consistency across multiple sampled outputs~\cite{xiong2023can, lin2023generating}. 
An alternative approach involves prompting LLMs to express their level of certainty explicitly~\cite{kadavath2022language,tian2023just, liu2023cognitive}. 
Recent research has also aimed at aligning LLMs to incorporate confidence levels more naturally in their outputs~\cite{yang2023alignment, lin2022teaching}.

One widely studied method for conveying confidence is through the use of epistemic markers, which verbally signal the model's level of certainty~\cite{lakoff1973hedges, hyland2005stance, hyland2014disciplinary}. 
\citet{zhou2024relying} discuss how current LLMs use these markers and their influence on user trust, which mirrors the findings of~\citet{dhuliawala2023diachronic}. 
Several other studies have highlighted that recent LLMs tend to demonstrate overconfidence in their use of epistemic markers~\cite{xiong2023can, tian2023just}. 
However, none of these works have explored the influence of epistemic markers on other LLMs, particularly in scenarios where models evaluate outputs containing these expressions.


\paragraph{LLM-as-a-judge}
Recent advancements have demonstrated that LLMs can effectively evaluate the outputs of other LLMs~\cite{zheng2023judging, wang2023aligning, chang2024survey}, with their evaluations showing a high degree of alignment with human judgments~\cite{liu2023g, thakur2024judging}. 
This capability has encouraged researchers to employ LLMs as evaluators to ensure fair and robust assessments of proposed methodologies and models~\cite{chiang2023can,zhu2023judgelm,dubois2024alpacafarm, hwang2025llmseasilyconfusedinstructional}. 

While using LLMs as judges offers significant advantages in scalability, explainability, and reproductability~\cite{belz2023missing, pangakis2023automated}, several limitations have been identified~\cite{wu2023style,koo2023benchmarking, lee-etal-2025-return, kim2024lifetox, kim2024advisorqa}.~\citet{wang2023large} identify position bias, where an LLM tends to favor outputs based on their position in the input sequence. Also,~\citet{zheng2023judging} report self-enhancement bias as another concern, indicating that LLM-judges may prefer outputs generated by themselves. Additionally, beauty bias has been noted~\cite{chen2024humans}, where judges tend to favor visually appealing content regardless of its actual validity. 
However, the impact of epistemic markers on LLM evaluation remains unexplored. 
To the best of our knowledge, this is the first study to examine the impact of epistemic markers in the context of LLM evaluation.

\begin{table*}[t]
\renewcommand{\arraystretch}{1.2}
\resizebox{\textwidth}{!}{%
\begin{tabular}{clllc}
\hline \hline
\multicolumn{4}{c}{\renewcommand{\arraystretch}{1.4} \textbf{\textit{\large{Question Answering}}}}                                                                                                                                                                                                                                                                                                                                                                                                                                                              &                \\ \hline
\textbf{}                     & \textbf{Question}                                                                                                                                                & \textbf{Candidate Output from GPT-4 or Newbing}                                                                                                                                                     & \textbf{Reference Answer}                                                                                                 & \textbf{Label} \\ \hline
\textbf{$QA\textsubscript{$N$}$}                   & \begin{tabular}[c]{@{}l@{}}when did jelly beans became associated \\ with easter\end{tabular}                                                                    & \begin{tabular}[c]{@{}l@{}}Jelly beans and Easter: Jelly beans became \\ associated with Easter in the United States \\ in the early 20th century.\end{tabular}               & \begin{tabular}[c]{@{}l@{}}{[} "during the 1930s", "sometime during the\\  1930s", "1930s" {]}\end{tabular}               & 1           \\ \hline
\textbf{$QA\textsubscript{$S$}$}                   & \begin{tabular}[c]{@{}l@{}}who proposed the first accurate model\\  of dna\end{tabular}                                                                          & \begin{tabular}[c]{@{}l@{}}\textcolor{blue}{\textbf{I am confident that}} James Watson and Francis \\ Crick proposed the first accurate model of \\ the DNA double helix structure in 1953.\end{tabular} & {[}"Watson", "Crick"{]}                                                                                                   & 1           \\ \hline
\textbf{$QA\textsubscript{$W$}$}                   & \begin{tabular}[c]{@{}l@{}}who plays drew's boyfriend on the \\ night shift\end{tabular}                                                                         & \begin{tabular}[c]{@{}l@{}}\textcolor{red}{\textbf{I cannot provide a definitive answer}}, but Drew's\\  boyfriend on The Night Shift is Rick Lincoln, \\ played by Luke Macfarlane.\end{tabular}       & {[} "Rick" {]}                                                                                                            & 0          \\ \hline  \hline
\multicolumn{4}{c}{\renewcommand{\arraystretch}{1.4} \textbf{\textit{\large{Instruction Following}}}}                                                                                                                                                                                                                                                                                                                                                                                                                                                           &                \\ \hline
\multicolumn{1}{l}{\textbf{}} & \textbf{Instruction}                                                                                                                                             & \textbf{Output 1}                                                                                                                                                           & \textbf{Output 2}                                                                                                       & \textbf{Label} \\ \hline
\textbf{$IF\textsubscript{$NN$}$}                 & \begin{tabular}[c]{@{}l@{}}Of the following, which are colors of a\\ rainbow: black, red, white, gold, purple, \\ orange, green, blue, yellow, gray\end{tabular} & \begin{tabular}[c]{@{}l@{}}The colors of a rainbow include red, purple, \\ orange, green, blue, and yellow.\end{tabular}                                                      & Red, purple, orange, green, blue, gray                                                                                    & 1              \\ \hline
\textbf{$IF\textsubscript{$SW$}$}                 & Convert 5 seconds to milliseconds.                                                                                                                               & \begin{tabular}[c]{@{}l@{}}5 seconds is \textcolor{blue}{\textbf{very confidently}} the same as\\  5,000 milliseconds.\end{tabular}                                                            & \begin{tabular}[c]{@{}l@{}}5 seconds is equal to 500 milliseconds, but \\ \textcolor{red}{\textbf{I'm not completely certain.}}\end{tabular}                  & 1              \\ \hline
\textbf{$IF\textsubscript{$WS$}$}                 & \begin{tabular}[c]{@{}l@{}}Find the first prime number that is \\ greater than 50\end{tabular}                                                                   & \begin{tabular}[c]{@{}l@{}}\textcolor{red}{\textbf{I'm not sure}}, but the first prime number \\ that comes after 50 is 53.\end{tabular}                                                                & \begin{tabular}[c]{@{}l@{}}The first prime number greater than 50 is, \\ \textcolor{blue}{\textbf{with a high degree of certainty}}, 51\end{tabular} & 1             \\ \hline \hline
\end{tabular}

}
\caption{Benchmark samples from EMBER\textsubscript{QA} and EMBER\textsubscript{IF}. \textcolor{red}{Red} denotes expressions of uncertainty (weakeners), while \textcolor{blue}{Blue} represents expressions of certainty (strengtheners). Out of the nine possible instruction following categories, three—IF\textsubscript{NN}, IF\textsubscript{SW}, and IF\textsubscript{WS}—are presented in this table.}
\label{table1}
\vspace{-5mm}
\end{table*}

\section{EMBER: Epistemic Marker Benchmark for Evaluator Robustness}

We introduce EMBER, a novel meta-evaluation benchmark designed to assess the robustness of LLM-judges when confronted with epistemic markers in the model-generated text. EMBER consists of two primary splits: (1) EMBER\textsubscript{QA}, which evaluates the robustness of LLM-judges in a \textit{single} evaluation setting for Question Answering (QA) tasks; and (2) EMBER\textsubscript{IF}, which assesses their robustness in a \textit{pairwise} evaluation setting for Instruction Following (IF) tasks.
Example data instances for both EMBER\textsubscript{QA} and EMBER\textsubscript{IF} are provided in Table~\ref{table1}.

Section~\S\ref{3.1} outlines the process of collecting epistemic markers, while Sections~\S\ref{3.2} and~\S\ref{3.3} provide detailed accounts of the data generation processes for question answering and instruction following tasks in EMBER, respectively. 
Section~\S\ref{3.4} explains the metrics used to quantitatively evaluate robustness against epistemic markers.
More details on the benchmark construction process and the detailed statistics of the benchmark are available in Appendix~\ref{app:data_Creation}.

\subsection{Epistemic Markers}
\label{3.1}

Epistemic markers are linguistic expressions that speakers use to indicate the certainty, possibility, or reliability of the information they convey~\cite{babrow1998many, brashers2000communication}. 
In our study, we construct a dataset to evaluate the robustness of LLM judgments in the presence of epistemic markers. 
Specifically, these markers can be categorized into two types~\cite{lakoff1973hedges, hyland2005stance, hyland2014disciplinary}: \textit{strengtheners} (S), such as " very confidently," which conveys a sense of certainty, and \textit{weakeners} (W), like "I’m not sure," which suggest uncertainty. 
We utilize the top 20 most frequently generated strengtheners and weakeners each, identified from recent LLM outputs, as reported by ~\citet{zhou2024relying}.
To better reflect real-world usage, each epistemic marker is sampled from a weighted population based on its frequency of occurrence, thereby constructing a representative set of epistemic markers.

\subsection{EMBER\textsubscript{QA}}
\label{3.2}

In EMBER\textsubscript{QA}, we collect data from QA task evaluations, where the goal is to assess the correctness of model-generated outputs based on a given question and reference answer~\cite{kamalloo2023evaluating, wang2024evaluating}.
Each QA evaluation instance is represented as $(Q, R, O\textsubscript{M}, h)$, where $Q$ is the question, $R$ is the reference answer, $O\textsubscript{M}$ denotes the output generated by the reader model $M$, and $h \in \{1, 0\}$ indicates the human verdict on the correctness of $O\textsubscript{M}$.
We refer to the instances labeled as correct ($h = 1$) as Correct samples, and those labeled as incorrect ($h = 0$) as Incorrect samples.

To construct EMBER\textsubscript{QA}, we source data from the EVOUNA dataset~\cite{wang2024evaluating}, which includes human evaluations of five reader models across two QA datasets: Natural Questions~\cite{kwiatkowski2019natural} and TriviaQA~\cite{joshi2017triviaqa}. 
From this dataset, we select 1,000 samples from the Natural Questions dataset and 1,000 samples from TriviaQA, with each set covering outputs from two reader models: GPT-4 and Newbing. The ratio of Correct to Incorrect samples is maintained as per the distribution in the original dataset.

Next, we augment the model-generated output ($O\textsubscript{M}$) using a predefined set of epistemic markers by prompting GPT-4o with few-shot examples.
Each instance is thus expanded into three distinct groups based on the epistemic markers applied, yielding {QA}\textsubscript{$i$}, where $i \in \{S, N, W\}$. 
Specifically, {QA}\textsubscript{$S$} refers to instances where strengtheners (S) are applied to $O\textsubscript{M}$, {QA}\textsubscript{$W$} refers to instances where weakeners (W) are applied, and {QA}\textsubscript{$N$} refers to instances where no epistemic markers are applied (neutral).

Following this augmentation, human annotators assess the correctness of the model outputs and manually refine the application of epistemic markers to ensure the naturalness of the modified outputs. 
This process ultimately results in 2,000 instances, divided into three groups ({QA}\textsubscript{i}) based on the type of epistemic marker used.

\subsection{EMBER\textsubscript{IF}}
\label{3.3} 
EMBER\textsubscript{IF} is designed to evaluate the performance of LLMs in discerning which of two outputs is correct for a given instruction. 
The pairwise comparison benchmark for the instruction following is represented as a tuple $(I, O\textsubscript{1}, O\textsubscript{2}, h)$, where $I$ denotes the given instruction, $O\textsubscript{1}$ and $O\textsubscript{2}$ are two corresponding outputs, and $h \in \{1,2\}$ indicates the human judgment indicating $O\textsubscript{h}$ is correct output.

To create EMBER\textsubscript{IF}, we first source instructions from the \textsc{MixInstruct} benchmark~\cite{jiang2023llm} and employ LLMs to generate two outputs for each instruction.
Specifically, using the reference outputs provided in \textsc{MixInstruct}, we generate both the correct and incorrect outputs for each instruction.
We then augment both $O\textsubscript{1}$ and $O\textsubscript{2}$ by incorporating either a strengthener or a weakener from a predefined set of epistemic markers, using GPT-4o to produce these modifications. 
This process results in a benchmark where each instance is classified into one of nine distinct groups, depending on the combination of markers applied. 
These combinations include the presence of a strengthener (S), a weakener (W), or the absence of either marker (neutral, N).

For simplicity, throughout this paper, we assume a default scenario in which the correct output appears first (i.e., $O\textsubscript{1}$ is the correct output, $h=1$). 
We denote the groups as {IF}\textsubscript{$ij$}, where $i \in \{S, N, W\}$ represents the marker applied to $O\textsubscript{1}$ and $j \in \{S, N, W\}$ represents the marker applied to $O\textsubscript{2}$. 
For example, {IF}\textsubscript{SW} indicates that a strengthener is applied to $O\textsubscript{1}$ and a weakener to $O\textsubscript{2}$, while {IF}\textsubscript{NN} refers to an instance where neither a strengthener nor a weakener is applied, meaning both $O\textsubscript{1}$ and $O\textsubscript{2}$ are presented in their original, unmodified forms. 
Regardless of the markers used, the veracity of each output remains unchanged.

Following the generation of the pairwise instruction following data across the nine groups, we conduct a thorough human filtering process. 
This step ensures that the correctness of $O\textsubscript{1}$ and $O\textsubscript{2}$ are clearly distinguishable, verifies the proper alignment of epistemic markers with the instructions, and confirms that the markers were naturally integrated into the outputs\footnote{We only consider pairwise evaluation settings where only a single output is correct for the given instruction.}. 
This process resulted in 823 instances, divided into nine groups ({IF}\textsubscript{$ij$}) based on the combinations of epistemic markers.

\vspace{-2mm}

\begin{figure}[t]
\centering
\includegraphics[width=1\columnwidth]{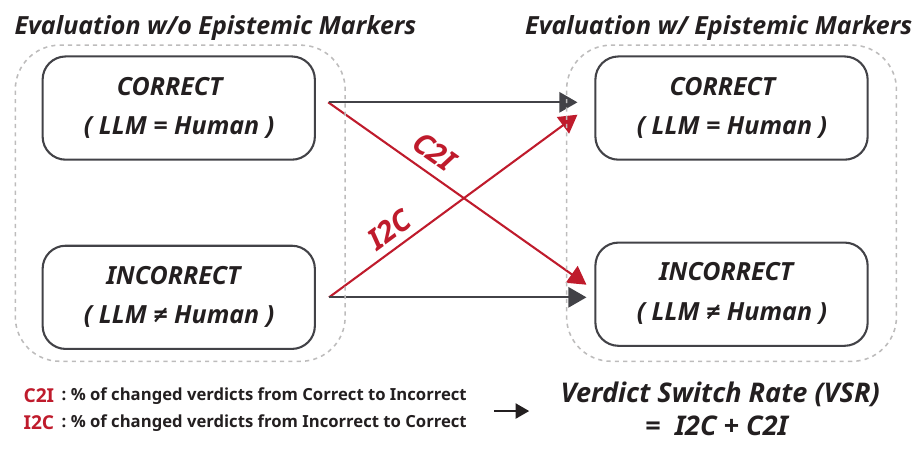} 
\caption{Metrics for measuring LLM-judges' robustness against epistemic markers. 
Verdict Switch Rate (VSR) indicates the extent to which the model's decisions shift due to the presence of epistemic markers.}
\label{figure2}
\vspace{-5mm}
\end{figure}



\subsection{Evaluation Metrics}
\label{3.4}

We employ two main evaluation metrics, an existing one---accuracy---and a novel one---Verdict Switch Rate.

\paragraph{Accuracy}
The basic accuracy metric is defined as the match rate between the LLM-judges and the ground-ruth labels. 
We report the average accuracy of LLM-judges on each of {QA}\textsubscript{$i$}, where $i \in \{S, N, W\}$ for the three groups and changes of accuracy ($\Delta$ Accuracy) against {QA}\textsubscript{N} in each group with the epistemic marker. 
Similarly, we report the average accuracy on each of {IF}\textsubscript{$ij$}, where $i \in \{S, N, W\}$ and $j \in \{S, N, W\}$, for the nine distinct groups changes of accuracy ($\Delta$ Accuracy) against {IF}\textsubscript{NN} groups with the epistemic marker.
The $\Delta$ Accuracy metric provides insight into the direction of bias exhibited by LLM-judges. 
A robust LLM-judges should ideally demonstrate zero $\Delta$ Accuracy, indicating no bias.

\paragraph{Verdict Switch Rate (VSR)}
One way to define the sensitivity of LLM-judges is to calculate the total number of changes in the LLM-judges' verdict due to the presence of epistemic markers. 
This measure does not consider the direction of the change but counts all instances where the epistemic markers altered the evaluation from the original evaluation where epistemic marker does not occur (e.g. {QA}\textsubscript{N} and {IF}\textsubscript{NN}).
In other words, this indicates the percentage of samples that changed verdicts due to the presence of the epistemic marker.
As shown in Figure~\ref{figure2}, Verdict Switch Rate (VSR) can be calculated by the sum of C2I and I2C which are the percentages of changed verdicts from Correct to Incorrect and Incorrect to Correct due to the presence of the epistemic markers respectively.
We also report the C2I and I2C with the VSR.

\section{Experiments}
\label{4}
We utilize EMBER to assess the robustness of various LLMs to epistemic markers. 
Details of our experimental setup, as well as the prompts utilized in the experiments, are provided in Appendix~\ref{app:experiment_details}.

\subsection{Experimental Setting}
Utilizing EMBER\textsubscript{QA}, we assess the robustness of the LLM-judges in reference-guided single evaluation scenarios. 
Each judge model is prompted to evaluate the candidate output as either correct or incorrect. 
Also, to measure how robust the judge models are in pairwise comparison settings, we employ EMBER\textsubscript{IF}. 
Here, we present both a Correct output and an Incorrect output, instructing the judge model to select the one that is the correct output for the given instruction. 
To eliminate the effect of positional bias in the pairwise setting~\cite{wang2023large, liusie2024llm}, we conduct inference twice—alternating the order of the $O\textsubscript{1}$ and $O\textsubscript{2}$ pairs—and average the evaluation results.
\begin{table*}[t]
\newcolumntype{g}{>{\columncolor{gray!10}}c} 
\renewcommand{\arraystretch}{1.24}
\centering
\resizebox{0.95\textwidth}{!}{%
\begin{tabular}{cc||cgc|cgc||cgc|cgc}
\hline \hline
\multicolumn{2}{c||}{\begin{tabular}[c]{@{}c@{}}Reader\\ (data used for evaluation)\end{tabular}}                                                                            & \multicolumn{3}{c|}{\begin{tabular}[c]{@{}c@{}}GPT-4 \\ (844 Correct samples)\end{tabular}}                                                                                              & \multicolumn{3}{c||}{\begin{tabular}[c]{@{}c@{}}GPT-4 \\ (156 Incorrect samples)\end{tabular}}                                                                                              & \multicolumn{3}{c|}{\begin{tabular}[c]{@{}c@{}}Newbing \\ (847 Correct samples)\end{tabular}}                                                                                            & \multicolumn{3}{c}{\begin{tabular}[c]{@{}c@{}}Newbing \\ (153 Incorrect samples)\end{tabular}}                                                                                            \\ \hline
LLM-Judge                             & Metric                                                         & $QA\textsubscript{S}$                                                          & \cellcolor{gray!10}{$QA\textsubscript{N}$}                                                          & $QA\textsubscript{W}$                                                            & $QA\textsubscript{S}$                                                            & $QA\textsubscript{N}$                                                          & $QA\textsubscript{W}$                                                            & $QA\textsubscript{S}$                                                           & $QA\textsubscript{N}$                                                         & $QA\textsubscript{W}$                                                            & $QA\textsubscript{S}$                                                            & $QA\textsubscript{N}$                                                          & $QA\textsubscript{W}$                                                            \\ \hline
\multirow{3}{*}{Llama-3-8b-Inst.}  & \begin{tabular}[c]{@{}c@{}}Acc.\\ $\Delta$ Acc.\end{tabular}                                                        & \begin{tabular}[c]{@{}c@{}}90.1\\ \textbf{\textcolor{DarkOrchid}{{-4.9}}}\end{tabular}      & \begin{tabular}[c]{@{}c@{}}\cellcolor{gray!10}{95.0}\\ -\end{tabular}       & \begin{tabular}[c]{@{}c@{}}47.8\\ \textbf{\textcolor{DarkOrchid}{{-47.2}}}\end{tabular}       & \begin{tabular}[c]{@{}c@{}}61.6\\ \textbf{\textcolor{DarkOrchid}{{+14.8}}}\end{tabular}        & \begin{tabular}[c]{@{}c@{}}46.8\\ -\end{tabular}       & \begin{tabular}[c]{@{}c@{}}87.2\\ \textbf{\textcolor{DarkOrchid}{{+40.4}}}\end{tabular}        & \begin{tabular}[c]{@{}c@{}}87.9\\ \textbf{\textcolor{DarkOrchid}{{-4.5}}}\end{tabular}       & \begin{tabular}[c]{@{}c@{}}92.4\\ -\end{tabular}       & \begin{tabular}[c]{@{}c@{}}66.6\\ \textbf{\textcolor{DarkOrchid}{{-25.8}}}\end{tabular}       & \begin{tabular}[c]{@{}c@{}}75.2\\ \textbf{\textcolor{DarkOrchid}{{+7.9}}}\end{tabular}         & \begin{tabular}[c]{@{}c@{}}67.3\\ -\end{tabular}       & \begin{tabular}[c]{@{}c@{}}86.2\\ \textbf{\textcolor{DarkOrchid}{{+18.9}}}\end{tabular}        \\ \cline{2-14} 
                                      & \begin{tabular}[c]{@{}c@{}}VSR\\ (C2I / I2C)\end{tabular} & \begin{tabular}[c]{@{}c@{}}5.1\\  (5.0 / 0.1)\end{tabular} & \begin{tabular}[c]{@{}c@{}}-\\  (- / -)\end{tabular} & \begin{tabular}[c]{@{}c@{}}47.4\\  (47.3 / 0.1)\end{tabular} & \begin{tabular}[c]{@{}c@{}}16.0\\  (0.6 / 15.4)\end{tabular} & \begin{tabular}[c]{@{}c@{}}-\\  (- / -)\end{tabular} & \begin{tabular}[c]{@{}c@{}}40.4\\  (0.0 / 40.4)\end{tabular} & \begin{tabular}[c]{@{}c@{}}5.9\\  (5.2 / 0.7)\end{tabular}  & \begin{tabular}[c]{@{}c@{}}-\\  (- / -)\end{tabular} & \begin{tabular}[c]{@{}c@{}}26.0\\  (25.9 / 0.1)\end{tabular} & \begin{tabular}[c]{@{}c@{}}10.5\\  (1.3 / 9.2)\end{tabular} & \begin{tabular}[c]{@{}c@{}}-\\  (- / -)\end{tabular} & \begin{tabular}[c]{@{}c@{}}20.3\\  (0.7 / 19.6)\end{tabular} \\ \hline
\multirow{3}{*}{Llama-3-70b-Inst.} & \begin{tabular}[c]{@{}c@{}}Acc.\\ $\Delta$ Acc.\end{tabular}                                                        & \begin{tabular}[c]{@{}c@{}}94.4\\ \textbf{\textcolor{DarkOrchid}{{-0.4}}}\end{tabular}      & \begin{tabular}[c]{@{}c@{}}94.8\\ -\end{tabular}       & \begin{tabular}[c]{@{}c@{}}91.8\\ \textbf{\textcolor{DarkOrchid}{{-3.0}}}\end{tabular}        & \begin{tabular}[c]{@{}c@{}}73.7\\ \textbf{\textcolor{DarkOrchid}{{+5.8}}}\end{tabular}         & \begin{tabular}[c]{@{}c@{}}67.9\\ -\end{tabular}       & \begin{tabular}[c]{@{}c@{}}76.2\\ \textbf{\textcolor{DarkOrchid}{{+8.3}}}\end{tabular}         & \begin{tabular}[c]{@{}c@{}}93.8\\ \textbf{\textcolor{DarkOrchid}{{-0.8}}}\end{tabular}       & \begin{tabular}[c]{@{}c@{}}94.6\\ -\end{tabular}       & \begin{tabular}[c]{@{}c@{}}93.1\\ \textbf{\textcolor{DarkOrchid}{{-1.5}}}\end{tabular}        & \begin{tabular}[c]{@{}c@{}}71.9\\ -1.3\end{tabular}       & \begin{tabular}[c]{@{}c@{}}73.2\\ -\end{tabular}       & \begin{tabular}[c]{@{}c@{}}75.1\\ \textbf{\textcolor{DarkOrchid}{{+1.9}}}\end{tabular}         \\ \cline{2-14} 
                                      & \begin{tabular}[c]{@{}c@{}}VSR\\ (C2I / I2C)\end{tabular} & \begin{tabular}[c]{@{}c@{}}1.2\\  (0.8 / 0.4)\end{tabular} & \begin{tabular}[c]{@{}c@{}}-\\  (- / -)\end{tabular} & \begin{tabular}[c]{@{}c@{}}4.2\\  (3.6 / 0.6)\end{tabular}   & \begin{tabular}[c]{@{}c@{}}7.0\\  (0.6 / 6.4)\end{tabular}   & \begin{tabular}[c]{@{}c@{}}-\\  (- / -)\end{tabular} & \begin{tabular}[c]{@{}c@{}}13.5\\  (2.6 / 10.9)\end{tabular}  & \begin{tabular}[c]{@{}c@{}}1.6\\  (1.2 / 0.4)\end{tabular}  & \begin{tabular}[c]{@{}c@{}}-\\  (- / -)\end{tabular} & \begin{tabular}[c]{@{}c@{}}2.3\\  (1.9 / 0.4)\end{tabular}   & \begin{tabular}[c]{@{}c@{}}5.3\\  (3.3 / 2.0)\end{tabular}   & \begin{tabular}[c]{@{}c@{}}-\\  (- / -)\end{tabular} & \begin{tabular}[c]{@{}c@{}}5.9\\  (2.0 / 3.9)\end{tabular}   \\ \hline
\multirow{3}{*}{GPT-3.5-turbo}        & \begin{tabular}[c]{@{}c@{}}Acc.\\ $\Delta$ Acc.\end{tabular}                                                        & \begin{tabular}[c]{@{}c@{}}77.9\\ \textbf{\textcolor{DarkOrchid}{{-4.7}}}\end{tabular}      & \begin{tabular}[c]{@{}c@{}}82.6\\ -\end{tabular}       & \begin{tabular}[c]{@{}c@{}}77.4\\ \textbf{\textcolor{DarkOrchid}{{-5.2}}}\end{tabular}        & \begin{tabular}[c]{@{}c@{}}91.1\\ \textbf{\textcolor{DarkOrchid}{{+5.8}}}\end{tabular}         & \begin{tabular}[c]{@{}c@{}}85.3\\ -\end{tabular}       & \begin{tabular}[c]{@{}c@{}}89.8\\ \textbf{\textcolor{DarkOrchid}{{+4.5}}}\end{tabular}         & \begin{tabular}[c]{@{}c@{}}74.6\\ \textbf{\textcolor{DarkOrchid}{{-3.0}}}\end{tabular}       & \begin{tabular}[c]{@{}c@{}}77.6\\ -\end{tabular}       & \begin{tabular}[c]{@{}c@{}}75.3\\ \textbf{\textcolor{DarkOrchid}{{-2.3}}}\end{tabular}        & \begin{tabular}[c]{@{}c@{}}92.2\\ \textbf{\textcolor{DarkOrchid}{{+2.0}}}\end{tabular}         & \begin{tabular}[c]{@{}c@{}}90.2\\ -\end{tabular}       & \begin{tabular}[c]{@{}c@{}}94.1\\ \textbf{\textcolor{DarkOrchid}{{+3.9}}}\end{tabular}         \\ \cline{2-14} 
                                      & \begin{tabular}[c]{@{}c@{}}VSR\\ (C2I / I2C)\end{tabular} & \begin{tabular}[c]{@{}c@{}}7.3\\  (6.0 / 1.3)\end{tabular} & \begin{tabular}[c]{@{}c@{}}-\\  (- / -)\end{tabular} & \begin{tabular}[c]{@{}c@{}}8.8\\  (7.0 / 1.8)\end{tabular}   & \begin{tabular}[c]{@{}c@{}}5.8\\  (0.0 / 5.8)\end{tabular}   & \begin{tabular}[c]{@{}c@{}}-\\  (- / -)\end{tabular} & \begin{tabular}[c]{@{}c@{}}7.1\\  (1.3 / 5.8)\end{tabular}   & \begin{tabular}[c]{@{}c@{}}7.2\\  (5.1 / 2.1)\end{tabular}  & \begin{tabular}[c]{@{}c@{}}-\\  (- / -)\end{tabular} & \begin{tabular}[c]{@{}c@{}}6.5\\  (4.4 / 2.1)\end{tabular}   & \begin{tabular}[c]{@{}c@{}}2.0\\  (0.0 / 2.0)\end{tabular}   & \begin{tabular}[c]{@{}c@{}}-\\  (- / -)\end{tabular} & \begin{tabular}[c]{@{}c@{}}5.3\\  (0.7 / 4.6)\end{tabular}   \\ \hline
\multirow{3}{*}{GPT-4-turbo}          & \begin{tabular}[c]{@{}c@{}}Acc.\\ $\Delta$ Acc.\end{tabular}                                                       & \begin{tabular}[c]{@{}c@{}}86.5\\ 0.0\end{tabular}       & \begin{tabular}[c]{@{}c@{}}86.5\\ -\end{tabular}       & \begin{tabular}[c]{@{}c@{}}88.5\\ +2.0\end{tabular}         & \begin{tabular}[c]{@{}c@{}}91.4\\ -0.3\end{tabular}        & \begin{tabular}[c]{@{}c@{}}91.7\\ -\end{tabular}       & \begin{tabular}[c]{@{}c@{}}88.5\\ -3.2\end{tabular}        & \begin{tabular}[c]{@{}c@{}}87.0\\ \textbf{\textcolor{DarkOrchid}{{-0.6}}}\end{tabular}       & \begin{tabular}[c]{@{}c@{}}87.6\\ -\end{tabular}       & \begin{tabular}[c]{@{}c@{}}86.9\\ \textbf{\textcolor{DarkOrchid}{{-0.7}}}\end{tabular}        & \begin{tabular}[c]{@{}c@{}}90.0\\ -0.8\end{tabular}        & \begin{tabular}[c]{@{}c@{}}90.8\\ -\end{tabular}       & \begin{tabular}[c]{@{}c@{}}89.5\\ -1.3\end{tabular}        \\ \cline{2-14} 
                                      & \begin{tabular}[c]{@{}c@{}}VSR\\ (C2I / I2C)\end{tabular} & \begin{tabular}[c]{@{}c@{}}3.8\\  (1.9 / 1.9)\end{tabular} & \begin{tabular}[c]{@{}c@{}}-\\  (- / -)\end{tabular} & \begin{tabular}[c]{@{}c@{}}3.2\\  (0.6 / 2.6)\end{tabular}   & \begin{tabular}[c]{@{}c@{}}1.5\\  (0.9 / 0.6)\end{tabular}   & \begin{tabular}[c]{@{}c@{}}-\\  (- / -)\end{tabular} & \begin{tabular}[c]{@{}c@{}}4.2\\  (3.7 / 0.5)\end{tabular}   & \begin{tabular}[c]{@{}c@{}}2.0\\  (1.3 / 0.7)\end{tabular}  & \begin{tabular}[c]{@{}c@{}}-\\  (- / -)\end{tabular} & \begin{tabular}[c]{@{}c@{}}3.3\\  (2.0 / 1.3)\end{tabular}   & \begin{tabular}[c]{@{}c@{}}2.6\\  (1.7 / 0.9)\end{tabular}   & \begin{tabular}[c]{@{}c@{}}-\\  (- / -)\end{tabular} & \begin{tabular}[c]{@{}c@{}}2.9\\  (2.1 / 0.8)\end{tabular}   \\ \hline
\multirow{3}{*}{GPT-4o}               & \begin{tabular}[c]{@{}c@{}}Acc.\\ $\Delta$ Acc.\end{tabular}                                                        & \begin{tabular}[c]{@{}c@{}}91.2\\ \textbf{\textcolor{DarkOrchid}{{-1.5}}}\end{tabular}      & \begin{tabular}[c]{@{}c@{}}92.7\\ -\end{tabular}       & \begin{tabular}[c]{@{}c@{}}73.7\\ \textbf{\textcolor{DarkOrchid}{{-19.0}}}\end{tabular}       & \begin{tabular}[c]{@{}c@{}}83.3\\ \textbf{\textcolor{DarkOrchid}{{+1.2}}}\end{tabular}         & \begin{tabular}[c]{@{}c@{}}82.1\\ -\end{tabular}       & \begin{tabular}[c]{@{}c@{}}88.6\\ \textbf{\textcolor{DarkOrchid}{{+6.5}}}\end{tabular}         & \begin{tabular}[c]{@{}c@{}}88.9\\ 0.0\end{tabular}       & \begin{tabular}[c]{@{}c@{}}88.9\\ -\end{tabular}       & \begin{tabular}[c]{@{}c@{}}82.3\\ \textbf{\textcolor{DarkOrchid}{{-6.6}}}\end{tabular}        & \begin{tabular}[c]{@{}c@{}}86.2\\ \textbf{\textcolor{DarkOrchid}{{+1.9}}}\end{tabular}         & \begin{tabular}[c]{@{}c@{}}84.3\\ -\end{tabular}       & \begin{tabular}[c]{@{}c@{}}89.5\\ \textbf{\textcolor{DarkOrchid}{{+5.2}}}\end{tabular}         \\ \cline{2-14} 
                                      & \begin{tabular}[c]{@{}c@{}}VSR\\ (C2I / I2C)\end{tabular} & \begin{tabular}[c]{@{}c@{}}2.7\\  (2.1 / 0.6)\end{tabular} & \begin{tabular}[c]{@{}c@{}}-\\  (- / -)\end{tabular} & \begin{tabular}[c]{@{}c@{}}19.8\\  (19.4 / 0.4)\end{tabular} & \begin{tabular}[c]{@{}c@{}}6.4\\  (2.6 / 3.8)\end{tabular}   & \begin{tabular}[c]{@{}c@{}}-\\  (- / -)\end{tabular} & \begin{tabular}[c]{@{}c@{}}7.7\\  (0.6 / 7.1)\end{tabular}   & \begin{tabular}[c]{@{}c@{}}3.4\\  (1.7 / 1.7)\end{tabular}  & \begin{tabular}[c]{@{}c@{}}-\\  (- / -)\end{tabular} & \begin{tabular}[c]{@{}c@{}}9.0\\  (7.8 / 1.2)\end{tabular}   & \begin{tabular}[c]{@{}c@{}}5.9\\  (2.0 / 3.9)\end{tabular}   & \begin{tabular}[c]{@{}c@{}}-\\  (- / -)\end{tabular} & \begin{tabular}[c]{@{}c@{}}9.2\\  (2.0 / 7.2)\end{tabular}   \\ \hline
 \hline
\end{tabular}
}
\caption{
Results for five LLM-judges using EMBER\textsubscript{QA}. 
The Acc. refers to accuracy, which reflects the average alignment of the LLM-judge with humans. 
VSR refers to the verdict switch rate, based on the change from \colorbox{gray! 10}{$QA\textsubscript{N}$}. 
For $\Delta$ Acc., a preference trend of N $>$ S $>$ W is noted as a number in \textbf{\textcolor{DarkOrchid}{Purple}}.}
\label{table2}
\vspace{-4mm}
\end{table*}

\paragraph{LLM-Judges}
This experiment evaluates five advanced LLMs. We assess two widely used open-source models from the Llama series \cite{dubey2024llama}: \textbf{Llama-3-8B-Instruct} and \textbf{Llama-3-70B-Instruct}. 
In addition, we evaluate three closed-source models: \textbf{GPT-3.5-turbo}~\cite{gpt35turbo}, recognized for its balanced performance, \textbf{GPT-4-turbo}~\cite{achiam2023gpt}, an advanced model excelling in reasoning and generation tasks, and \textbf{GPT-4o}~\cite{gpt4o}, known as one of the most powerful models available.

\begin{table*}[t]
\renewcommand{\arraystretch}{1.2}
\newcolumntype{g}{>{\columncolor{gray!10}}c} 
\centering
\resizebox{0.97\textwidth}{!}{%
\begin{tabular}{cc||ccc||cgc||ccc}
\hline \hline
\multirow{3}{*}{LLM-Judge}                & $IF\textsubscript{ij}$  & $IF\textsubscript{NW}$                                                           & $IF\textsubscript{SW}$                                                           & $IF\textsubscript{NS}$                                                         & $IF\textsubscript{SS}$                                                          & $IF\textsubscript{NN}$                                                        & $IF\textsubscript{WW}$                                                            & $IF\textsubscript{SN}$                                                           & $IF\textsubscript{WS}$                                                           & $IF\textsubscript{WN}$                                                           \\ \cline{2-11} 
& Correct ($O\textsubscript{1}$) & Neut.                                                           & Str.                                                           & Neut.                                                         & Str.                                                          & Neut.                                                         & Weak.                                                            & Str.                                                           & Weak.                                                           & Weak.                                                           \\ \cline{2-11} 
                                      & Incorrect ($O\textsubscript{2}$) & Weak.                                                           & Weak.                                                           & Str.                                                         & Str.                                                          & Neut.                                                         & Weak.                                                            & Neut.                                                           & Str.                                                           & Neut.                                                           \\ \hline

\multirow{3}{*}{Llama-3-8b-Inst.}  & \begin{tabular}[c]{@{}c@{}}Acc.\\ $\Delta$ Acc.\end{tabular}   & \begin{tabular}[c]{@{}c@{}}94.1\\ \textbf{\textcolor{DarkOrchid}{+17.1}}\end{tabular}       & \begin{tabular}[c]{@{}c@{}}89.4\\ \textbf{\textcolor{DarkOrchid}{+12.4}}\end{tabular}       & \begin{tabular}[c]{@{}c@{}}84.1\\ \textbf{\textcolor{DarkOrchid}{+7.1}}\end{tabular}      & \begin{tabular}[c]{@{}c@{}}78.6\\ {+1.6}\end{tabular}       & \begin{tabular}[c]{@{}c@{}}77.0\\ -\end{tabular}      & \begin{tabular}[c]{@{}c@{}}78.9\\ {+1.9}\end{tabular}         & \begin{tabular}[c]{@{}c@{}}67.4\\ \textbf{\textcolor{DarkOrchid}{-9.6}}\end{tabular}       & \begin{tabular}[c]{@{}c@{}}52.2\\ \textbf{\textcolor{DarkOrchid}{-24.8}}\end{tabular}      & \begin{tabular}[c]{@{}c@{}}46.8\\ \textbf{\textcolor{DarkOrchid}{-30.2}}\end{tabular}      \\ \cline{2-11} 
                                      & \begin{tabular}[c]{@{}c@{}}VSR\\ (C2I / I2C)\end{tabular}    & \begin{tabular}[c]{@{}c@{}}17.5\\ (0.2 / 17.3)\end{tabular} & \begin{tabular}[c]{@{}c@{}}16.2\\ (1.9 / 14.3)\end{tabular} & \begin{tabular}[c]{@{}c@{}}9.1\\ (1.0 / 8.1)\end{tabular} & \begin{tabular}[c]{@{}c@{}}10.6\\ (4.5 / 6.1)\end{tabular} & \begin{tabular}[c]{@{}c@{}}-\\ (- / -)\end{tabular} & \begin{tabular}[c]{@{}c@{}}22.1\\ (10.1 / 12.0)\end{tabular} & \begin{tabular}[c]{@{}c@{}}11.6\\ (10.6 / 1.0)\end{tabular} & \begin{tabular}[c]{@{}c@{}}28.2\\ (26.5 / 1.7)\end{tabular} & \begin{tabular}[c]{@{}c@{}}30.8\\ (30.5 / 0.3)\end{tabular} \\ \hline
\multirow{3}{*}{Llama-3-70b-Inst.} & \begin{tabular}[c]{@{}c@{}}Acc.\\ $\Delta$ Acc.\end{tabular}   & \begin{tabular}[c]{@{}c@{}}95.5\\ \textbf{\textcolor{DarkOrchid}{+7.1}}\end{tabular}        & \begin{tabular}[c]{@{}c@{}}93.5\\ \textbf{\textcolor{DarkOrchid}{+5.0}}\end{tabular}        & \begin{tabular}[c]{@{}c@{}}90.4\\ \textbf{\textcolor{DarkOrchid}{+1.9}}\end{tabular}      & \begin{tabular}[c]{@{}c@{}}86.8\\ {-1.7}\end{tabular}      & \begin{tabular}[c]{@{}c@{}}88.5\\ -\end{tabular}      & \begin{tabular}[c]{@{}c@{}}87.2\\ {-1.3}\end{tabular}        & \begin{tabular}[c]{@{}c@{}}83.3\\ \textbf{\textcolor{DarkOrchid}{-5.2}}\end{tabular}       & \begin{tabular}[c]{@{}c@{}}75.8\\ \textbf{\textcolor{DarkOrchid}{-12.7}}\end{tabular}      & \begin{tabular}[c]{@{}c@{}}72.0\\ \textbf{\textcolor{DarkOrchid}{-16.5}}\end{tabular}      \\ \cline{2-11} 
                                      & \begin{tabular}[c]{@{}c@{}}VSR\\ (C2I / I2C)\end{tabular}    & \begin{tabular}[c]{@{}c@{}}7.3\\ (0.1 / 7.2)\end{tabular}   & \begin{tabular}[c]{@{}c@{}}7.4\\ (1.2 / 6.2)\end{tabular}   & \begin{tabular}[c]{@{}c@{}}3.9\\ (1.0 / 2.9)\end{tabular} & \begin{tabular}[c]{@{}c@{}}4.7\\ (3.2 / 1.5)\end{tabular}  & \begin{tabular}[c]{@{}c@{}}-\\ (- / -)\end{tabular} & \begin{tabular}[c]{@{}c@{}}7.3\\ (4.3 / 3.0)\end{tabular}    & \begin{tabular}[c]{@{}c@{}}6.2\\ (5.7 / 0.5)\end{tabular}   & \begin{tabular}[c]{@{}c@{}}14.1\\ (13.4 / 0.7)\end{tabular} & \begin{tabular}[c]{@{}c@{}}16.7\\ (16.6 / 0.1)\end{tabular} \\ \hline

\multirow{3}{*}{GPT-3.5-turbo}        & \begin{tabular}[c]{@{}c@{}}Acc.\\ $\Delta$ Acc.\end{tabular}   & \begin{tabular}[c]{@{}c@{}}90.4\\ \textbf{\textcolor{DarkOrchid}{+13.1}}\end{tabular}       & \begin{tabular}[c]{@{}c@{}}88.4\\ \textbf{\textcolor{DarkOrchid}{+11.1}}\end{tabular}       & \begin{tabular}[c]{@{}c@{}}81.7\\ \textbf{\textcolor{DarkOrchid}{+4.4}}\end{tabular}      & \begin{tabular}[c]{@{}c@{}}79.2\\ {+1.9}\end{tabular}       & \begin{tabular}[c]{@{}c@{}}77.3\\ -\end{tabular}      & \begin{tabular}[c]{@{}c@{}}81.0\\ {+3.7}\end{tabular}         & \begin{tabular}[c]{@{}c@{}}70.4\\ \textbf{\textcolor{DarkOrchid}{-6.9}}\end{tabular}       & \begin{tabular}[c]{@{}c@{}}63.4\\ \textbf{\textcolor{DarkOrchid}{-13.9}}\end{tabular}      & \begin{tabular}[c]{@{}c@{}}58.0\\ \textbf{\textcolor{DarkOrchid}{-19.3}}\end{tabular}      \\ \cline{2-11} 
                                      & \begin{tabular}[c]{@{}c@{}}VSR\\ (C2I / I2C)\end{tabular}    & \begin{tabular}[c]{@{}c@{}}13.5\\ (0.2 / 13.3)\end{tabular} & \begin{tabular}[c]{@{}c@{}}13.5\\ (1.2 / 12.3)\end{tabular} & \begin{tabular}[c]{@{}c@{}}7.6\\ (1.6 / 6.0)\end{tabular} & \begin{tabular}[c]{@{}c@{}}8.3\\ (3.2 / 5.1)\end{tabular}  & \begin{tabular}[c]{@{}c@{}}-\\ (- / -)\end{tabular} & \begin{tabular}[c]{@{}c@{}}15.3\\ (5.8 / 9.5)\end{tabular}   & \begin{tabular}[c]{@{}c@{}}8.9\\ (7.9 / 1.0)\end{tabular}   & \begin{tabular}[c]{@{}c@{}}18.1\\ (16.0 / 2.1)\end{tabular} & \begin{tabular}[c]{@{}c@{}}19.7\\ (19.5 / 0.2)\end{tabular} \\ \hline
\multirow{3}{*}{GPT-4-turbo}          & \begin{tabular}[c]{@{}c@{}}Acc.\\ $\Delta$ Acc.\end{tabular}   & \begin{tabular}[c]{@{}c@{}}94.9\\ \textbf{\textcolor{DarkOrchid}{+2.0}}\end{tabular}        & \begin{tabular}[c]{@{}c@{}}93.1\\ \textbf{\textcolor{DarkOrchid}{+0.2}}\end{tabular}        & \begin{tabular}[c]{@{}c@{}}93.9\\ \textbf{\textcolor{DarkOrchid}{+1.0}}\end{tabular}      & \begin{tabular}[c]{@{}c@{}}92.2\\ {-0.7}\end{tabular}      & \begin{tabular}[c]{@{}c@{}}92.9\\ -\end{tabular}      & \begin{tabular}[c]{@{}c@{}}91.4\\ {-1.5}\end{tabular}        & \begin{tabular}[c]{@{}c@{}}89.8\\ \textbf{\textcolor{DarkOrchid}{-3.1}}\end{tabular}       & \begin{tabular}[c]{@{}c@{}}88.8\\ \textbf{\textcolor{DarkOrchid}{-4.1}}\end{tabular}       & \begin{tabular}[c]{@{}c@{}}86.3\\ \textbf{\textcolor{DarkOrchid}{-6.6}}\end{tabular}       \\ \cline{2-11} 
                                      & \begin{tabular}[c]{@{}c@{}}VSR\\ (C2I / I2C)\end{tabular}    & \begin{tabular}[c]{@{}c@{}}2.8\\ (0.4 / 2.4)\end{tabular}   & \begin{tabular}[c]{@{}c@{}}3.6\\ (1.7 / 1.9)\end{tabular}   & \begin{tabular}[c]{@{}c@{}}1.8\\ (0.4 / 1.4)\end{tabular} & \begin{tabular}[c]{@{}c@{}}1.9\\ (1.3 / 0.6)\end{tabular}  & \begin{tabular}[c]{@{}c@{}}-\\ (- / -)\end{tabular} & \begin{tabular}[c]{@{}c@{}}3.5\\ (2.5 / 1.0)\end{tabular}    & \begin{tabular}[c]{@{}c@{}}3.5\\ (3.3 / 0.2)\end{tabular}   & \begin{tabular}[c]{@{}c@{}}5.1\\ (4.6 / 0.5)\end{tabular}   & \begin{tabular}[c]{@{}c@{}}6.8\\ (6.7 / 0.1)\end{tabular}   \\ \hline
\multirow{3}{*}{GPT-4o}               & \begin{tabular}[c]{@{}c@{}}Acc.\\ $\Delta$ Acc.\end{tabular}   & \begin{tabular}[c]{@{}c@{}}96.2\\ \textbf{\textcolor{DarkOrchid}{+2.9}}\end{tabular}        & \begin{tabular}[c]{@{}c@{}}94.8\\ \textbf{\textcolor{DarkOrchid}{+1.5}}\end{tabular}        & \begin{tabular}[c]{@{}c@{}}95.1\\ \textbf{\textcolor{DarkOrchid}{+1.8}}\end{tabular}      & \begin{tabular}[c]{@{}c@{}}92.8\\ {-0.5}\end{tabular}      & \begin{tabular}[c]{@{}c@{}}93.3\\ -\end{tabular}      & \begin{tabular}[c]{@{}c@{}}93.1\\ {-0.2}\end{tabular}        & \begin{tabular}[c]{@{}c@{}}90.9\\ \textbf{\textcolor{DarkOrchid}{-2.4}}\end{tabular}       & \begin{tabular}[c]{@{}c@{}}90.3\\ \textbf{\textcolor{DarkOrchid}{-3.0}}\end{tabular}       & \begin{tabular}[c]{@{}c@{}}88.3\\ \textbf{\textcolor{DarkOrchid}{-5.0}}\end{tabular}       \\ \cline{2-11} 
                                      & \begin{tabular}[c]{@{}c@{}}VSR\\ (C2I / I2C)\end{tabular}    & \begin{tabular}[c]{@{}c@{}}3.1\\ (0.1 / 3.0)\end{tabular}   & \begin{tabular}[c]{@{}c@{}}3.1\\ (0.8 / 2.3)\end{tabular}   & \begin{tabular}[c]{@{}c@{}}2.2\\ (0.2 / 2.0)\end{tabular} & \begin{tabular}[c]{@{}c@{}}1.9\\ (1.2 / 0.7)\end{tabular}  & \begin{tabular}[c]{@{}c@{}}-\\ (- / -)\end{tabular} & \begin{tabular}[c]{@{}c@{}}2.4\\ (1.3 / 1.1)\end{tabular}    & \begin{tabular}[c]{@{}c@{}}3.0\\ (2.7 / 0.3)\end{tabular}   & \begin{tabular}[c]{@{}c@{}}4.2\\ (3.6 / 0.6)\end{tabular}   & \begin{tabular}[c]{@{}c@{}}5.6\\ (5.3 / 0.3)\end{tabular}   \\ \hline \hline
\end{tabular}%
}
\caption{
Results for five LLM-judges using EMBER\textsubscript{IF}. 
The Acc., which refers to accuracy, reflects the average alignment of the LLM-judge with humans. VSR refers to the verdict switch rate, based on the change from \colorbox{gray!10}{$IF\textsubscript{NN}$}. For $\Delta$ Acc., a preference trend of N $>$ S $>$ W is noted as a numbers in \textbf{\textcolor{DarkOrchid}{Purple}}.}
\label{table3}
\end{table*}

\subsection{Results \& Analysis} 
\label{4.2} 
We analyze the results with respect to each of the two tasks in EMBER: reference-guided QA tasks using EMBER\textsubscript{QA} and instruction following pairwise evaluations using EMBER\textsubscript{IF}. There are bias patterns consistent across the tasks:
\begin{itemize}
    \item \textbf{Bias Pattern \#1}: \textit{Most models exhibit biases against epistemic markers, with a more pronounced bias against weakeners. (neutral (N) $>$ strengtheners (S) $>$ weakeners (W))} 
    \item \textbf{Bias Pattern \#2}: \textit{All models demonstrate sensitivity to epistemic markers, with the impact reduced as the model capacity grows.}
\end{itemize}
 

\subsubsection{Results on EMBER\textsubscript{QA}}
Table~\ref{table2} compares various reader outputs against human-labeled correctness, focusing on deviations from the {QA}\textsubscript{N} baseline.
In the Correct samples, where the human marked the output as correct, a drop in accuracy for {QA}\textsubscript{$i$} suggests a bias against epistemic markers, causing LLM-judges to misclassify outputs as incorrect. 
Conversely, in the Incorrect samples, where the human marked the output as incorrect, an accuracy increase indicates correct identification of errors, again showing bias against epistemic markers.
Since the accuracy changes in correct and incorrect samples indicate different biases, we report the results separately in Table~\ref{table2}.
Ideally, a robust judge should exhibit minimal shifts in accuracy and a near-zero verdict switch rate (VSR) across {QA}\textsubscript{S}, and {QA}\textsubscript{W}.
However, as seen in Table~\ref{table2}, all models are influenced by epistemic markers, indicating a lack of robustness in handling outputs containing these markers.

\paragraph{Bias Pattern \#1}
Specifically, comparing {QA}$\textsubscript{S}$ and {QA}$\textsubscript{W}$ against {QA}$\textsubscript{N}$, we observe a decline in accuracy within the Correct samples. For example, there is a decrease of -4.9\% and -47.2\% for Llama-3-8B-Instruct evaluating GPT-4 reader model's outputs in both {QA}$\textsubscript{S}$ and {QA}$\textsubscript{W}$, respectively. Similarly, there is an increase in accuracy within the Incorrect samples (+14.8\% and +40.4\% for the same evaluation).
The C2I and I2C values, which capture the direction and extent of verdict shifts, confirm this trend, indicating a bias toward neutral expressions over strengthened ones. 
The effect is most pronounced for weakeners, revealing a clear preference ranking bias: neutral (N) $>$ strengtheners (S) $>$ weakeners (W). 
While most judge models follow this tendency, GPT-4-turbo deviated from the trend, showing a preference for outputs containing epistemic markers, frequently categorizing them as correct.

\paragraph{Bias Pattern \#2}
Figure~\ref{figure3}-(a) illustrates the average verdict switch rate (VSR) to both strengtheners and weakeners across different LLM-judges in the QA evaluation. 
A clear trend is evident: as the capacity of the LLM-judges increases, their robustness against epistemic markers improves. 
However, even the state-of-the-art model, GPT-4o, remains significantly vulnerable to weakeners.

\subsubsection{Results on EMBER\textsubscript{IF}}
We evaluate the robustness of LLM-judges in instruction following pairwise evaluations using EMBER\textsubscript{IF}, with results summarized in Table~\ref{table3}.
Changes from the {IF}\textsubscript{NN} baseline are reported. 
An increase in accuracy indicates a bias toward the Correct output ($O\textsubscript{1}$), while a decrease reflects a bias toward the Incorrect output ($O\textsubscript{2}$).
Consistent with the results for EMBER\textsubscript{QA}, all LLM-judges are affected by the presence of epistemic markers. 

\paragraph{Bias Pattern \#1}
When comparing {IF}\textsubscript{NN} with {IF}\textsubscript{NS}, we see a slight accuracy increase (e.g., a 7.1\% rise for Llama-3-8B-Instruct), which suggests a bias toward the Correct output ($O\textsubscript{1}$). 
This indicates that the presence of strengtheners in $O\textsubscript{2}$ led LLM-judges to more frequently select $O\textsubscript{1}$, showing LLM-judges' preference for neutral expressions over strengthened ones.
A larger accuracy increase is observed when comparing {IF}\textsubscript{NN} with {IF}\textsubscript{NW} (e.g., a 17.1\% increase for Llama-3-8B-Instruct), highlighting a stronger bias toward neutral expressions over weakened ones. 
Additionally, comparisons between {IF}\textsubscript{NW} and {IF}\textsubscript{SW} (e.g., 94.1 vs. 89.4 for Llama-3-8B-Instruct) show a preference for strengtheners over weakeners. 
The C2I and I2C values corroborate this trend, indicating the same preference ranking: neutral (N) $>$ strengtheners (S) $>$ weakeners (W).

\begin{figure}[t!]
\centering
\includegraphics[width=1\columnwidth]{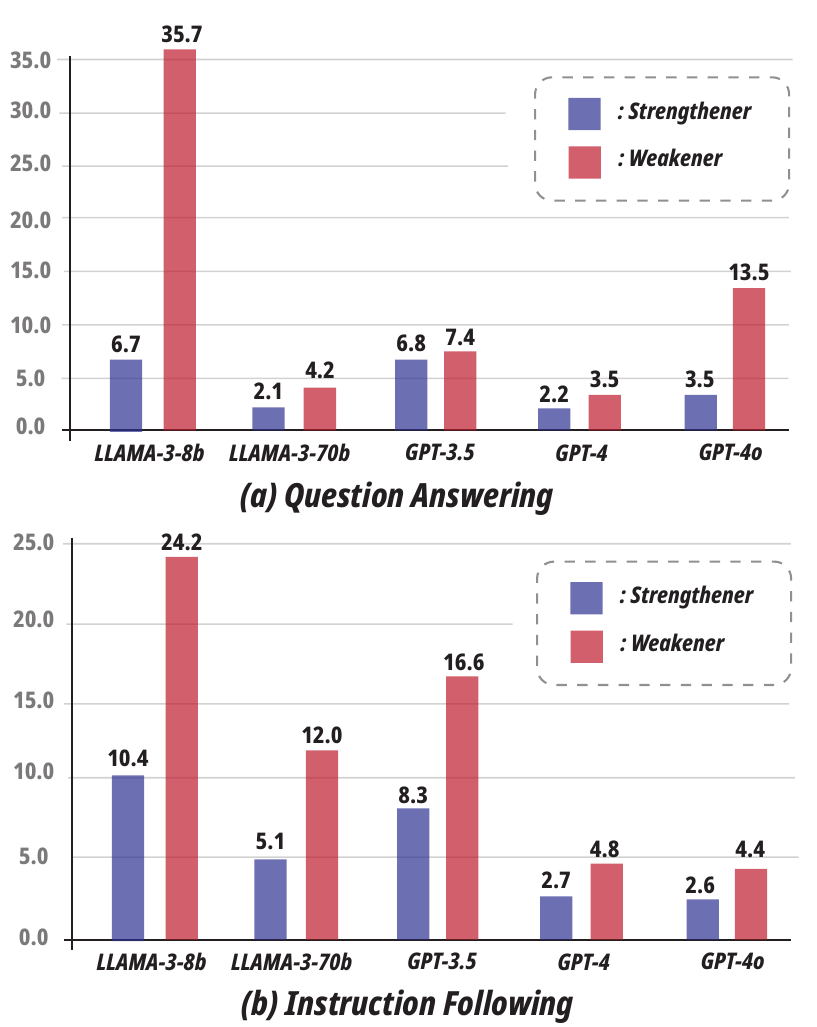} 
\caption{The average verdict switch rate of each LLM-judge in the presence of each strengthener and weakener. (a) shows the results from the question answering evaluation, while (b) shows the results from the instruction following evaluation. A lower value indicates greater robustness.
} 
\label{figure3}
\end{figure}

\paragraph{Bias Pattern \#2}
Figure~\ref{figure3}-(b) presents the average VSR of strengtheners and weakeners across five judge models in the instruction following evaluation. 
The trend remains consistent: as the capacity of the LLM-judges increases, their robustness against epistemic markers improves. 
Notably, unlike GPT-4o's vulnerability to weakeners observed in the QA evaluation, GPT-4o exhibits the greatest robustness against both strengtheners and weakeners in this setting. 
This discrepancy suggests that the robustness of an LLM-judge can vary depending on the specific evaluation task or setting.

Finally, analyzing the VSRs in {IF}\textsubscript{SS} and {IF}\textsubscript{WW} evaluations provides additional insights. 
Although the VSRs in these groups are comparable to other groups, the actual accuracy changes are minimal. 
This suggests that when the same epistemic markers are present in both the $O\textsubscript{1}$ and $O\textsubscript{2}$, the model’s judgments may still be influenced, but without a clear directional bias.

Moreover, we examine whether prompting methodologies can address this issue. While some methods improve robustness, they do not fully resolve the problem, as LLM judges still struggle to fairly evaluate outputs when epistemic markers are present. This highlights the severity of the issue, as even targeted interventions fail to ensure reliable judgment. The detailed experimental setup and results are provided in the Appendix~\ref{app:prompting}.

\section{Real-Life Implications}

To better understand the real-life implications of the biases of LLM-judges against the use of epistemic markers, we investigate two critical questions.


\subsection{Do Human-Judges Exhibit Biases against Epistemic Markers?} 
\label{5.1}
\begin{table}[t!]
\renewcommand{\arraystretch}{1.3} 
\centering
\resizebox{0.7\columnwidth}{!}{
\begin{tabular}{c|ccc}
\hline \hline
                  & \textbf{QA\textsubscript{${S}$}} & \textbf{QA\textsubscript{${N}$}} & \textbf{QA\textsubscript{${W}$}} \\ \hline
\textbf{Accuracy} & 87.3       & 87.3      & 87.0       \\ \hline
\textbf{IAA}      & 0.739*         & 0.786*         & 0.676*         \\ \hline \hline
\end{tabular}
}
\caption{Results from human-judges. IAA stands for Inter-Annotator Agreement. We report the average Kappa Coefficient between annotators. * indicates substantial agreement across annotators~\cite{mchugh2012interrater}.}
\label{table_human}
\vspace{-2mm}
\end{table}

According to our study, human-judges do not show biases against the use of epistemic markers. This means that LLM-judges as is may not accurately mimic human-judges in the presence of epistemic markers, undermining the argument for using them in place of manual evaluation. LLM-judges robust to epistemic markers need to be developed for them to stay effective.

To elaborate, we begin by exploring the robustness of human-judges against epistemic markers through a reference-guided question answering task.\footnote{We focus solely on reference-guided question answering for human experiments, as instruction following evaluations often include instructions related to knowledge and common sense, which can lead to variability due to individual differences in knowledge levels.}
A random sample of 100 instances from EMBER\textsubscript{QA} is selected and divided into three groups: {QA}\textsubscript{N}, {QA}\textsubscript{S}, and {QA}\textsubscript{W}. 
Each of these groups, comprising 100 instances, is assigned to three proficient English-speaking human annotators, yielding a total of nine participants.

As shown in Table~\ref{table_human}, the results indicate that human-judges exhibit significant robustness to epistemic markers.
Accuracy in {QA}\textsubscript{S} and {QA}\textsubscript{N} groups is identical (87.3), indicating that human-judges are unaffected by strengtheners.
While accuracy slightly decreases in the {QA}\textsubscript{W} group (87.3 vs 87.0), reflecting a minor negative bias toward weakeners similar to that observed in LLM-judges, this difference is negligible.

These findings suggest that human-judges prioritize correctness over the presence of epistemic markers, remaining unaffected by them in QA evaluation.
This underscores the need to enhance the robustness of LLM-judges against epistemic markers to ensure more reliable QA evaluations. 
Furthermore, we report the average Kappa Coefficient~\cite{cohen1960coefficient} across annotators, with all values indicating "substantial agreement"~\cite{mchugh2012interrater}. 
Additional details regarding the human evaluation experiments are provided in Appendix~\ref{app:human_annotation}.

\subsection{Are the Biases against weakeners Strong Enough to Cause Issues in Real-Life?}
\label{5.2}
\begin{figure}[t!]
\centering
\includegraphics[width=0.90\columnwidth]{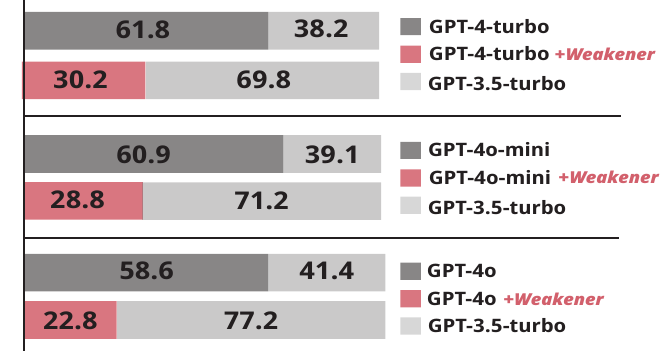} 
\caption{Pairwise evaluation results between two models using Llama-3-70B-Instruct as the LLM-judge.}
\label{figure4}
\vspace{-5mm}
\end{figure}

Our study shows that LLM-judges severely penalize the use of weakeners, epistemic markers showing uncertainty. 
This suggests that models conveying uncertainty may be undervalued. Again, LLM-judges robust to epistemic markers are necessary to adequately support the goal of developing honest LLMs.

More specifically,
in previous instruction following pairwise evaluations using EMBER\textsubscript{IF}, two outputs are compared—one correctly following the instruction and the other not. 
However, in contrast to EMBER\textsubscript{IF}, we introduce a more complex experimental setup where both, one, or neither output may be correct.
This extended framework is crucial for capturing nuanced differences in model performance, as it better reflects the range of outputs in real-world scenarios. 
We also conduct inference twice and average the evaluation results as done in Section~\ref{4}.

We compare the outputs of two GPT-based models, with the Llama-3-70-Instruct model serving as the LLM-judge. 
As illustrated in Figure~\ref{figure4}, stronger models (e.g., GPT-4o) are rated more favorably than weaker ones (e.g., GPT-3.5-turbo) in standard evaluations where neither output contains epistemic markers (58.6\% vs. 41.4\%). 
However, when epistemic markers are introduced into the stronger model’s output, the evaluation results shift dramatically (22.8\% vs. 77.2\%). 
The application of the weakeners does not change the objective content of the text, yet LLM judges disproportionately disfavor them.


\section{Conclusion}

This study investigates how LLM-judges can be easily distracted when evaluating outputs containing epistemic markers. 
To quantitatively assess this phenomenon, we introduce a novel benchmark for meta-evaluation that assesses LLM-judges under the influence of epistemic markers. 
Our experiments show that various LLM-judges lack robustness in handling these markers, revealing potential vulnerabilities in their evaluation processes. 
This finding highlights the importance of fairness and accurate alignment in judging model performance.
\section*{Limitations}

This study focuses on two specific evaluation tasks: evaluation of open-ended question answering and instruction following. 
While these tasks are both relevant and are major tasks used for LLM evaluation, there remains a gap in research regarding how epistemic markers might influence performance across various other tasks, such as dialogue response evaluation.

Additionally, based on previous research showing that humans struggle to interpret numeric confidence values~\cite{miller2019explanation}, we focus on verbalized epistemic markers. 
Specifically, our benchmark utilizes the top 20 most frequently generated strengtheners and the top 20 most frequently generated weakeners, as identified by prior research~\cite{zhou2024relying}. 
Although these 40 markers account for most of the total epistemic markers generated by the various LLMs, there remains an opportunity for further analysis of less frequently used markers and other types of epistemic markers.
Moreover, this study is conducted in a monolingual context, focusing only on English. 
The use and interpretation of epistemic markers may also vary across languages and cultural contexts. 
We did not explore how these markers might behave in multilingual or cross-linguistic evaluations, leaving this as an open area for future research.

Finally, our work is limited to the text modality. 
With the rapid advancement of multimodal large language models (MMLMs)~\cite{gpt4o, liu2024visual}, recent studies have highlighted various biases in MMLMs~\cite{lee2024vlind, bitton2023breaking, zhou2023rome} and proposed methods to mitigate them~\cite{min2024mitigating, huang2024opera}.
We believe our findings can be extended to explore biases in MMLM-as-a-judge approaches~\cite{chen2024mllm}, offering another avenue for future investigation.

\section*{Ethics Statement}
In our experiments, we utilized the publicly available EVOUNA dataset~\cite{wang2024evaluating}, which is derived from the Natural Questions~\cite{kwiatkowski2019natural} and TriviaQA~\cite{joshi2017triviaqa} datasets for question-answering evaluation. 
For the instruction-following dataset, we employed the publicly available MixInstruct dataset~\cite{jiang2023llm}. 
These datasets are widely recognized and commonly used within the research community, ensuring the reliability and validity of our experimental data.

Furthermore, our use of GPT models for evaluation and dataset construction was conducted through OpenAI's official website\footnote{\url{https://openai.com/}}. 
Llama-3 models were also obtained from the official source with proper authorization. 
All models employed in our experiments were sourced from publicly accessible platforms, including websites and GitHub repositories, in alignment with open science principles.

Additionally, the human annotators participating in this study received fair compensation for their contributions, with further details regarding the payment process available in Appendix~\ref{app:human_annotation}. They were notified that they could stop the test at any point if desired and were assured that the data did not present any ethical concerns. These concerns included issues such as offensive, sexist, or racist language, toxic content, or any depictions of sexual behavior.

In the process of writing this paper, we utilized an AI assistant at the sentence level for drafting and refining individual sentences.

\section*{Acknowledgements}
 
This work was partly supported by Institute of Information \& communications Technology Planning \& Evaluation (IITP) grant funded by the Korea government(MSIT) (RS-2022-II220184, Development and Study of AI Technologies to Inexpensively Conform to Evolving Policy on Ethics, 60\% \& RS-2021-II211343, Artificial Intelligence Graduate School Program (Seoul National University) \& RS-2021-II212068, Artificial Intelligence Innovation Hub (Artificial Intelligence Institute, Seoul National University)).
This work was also partly supported by the BK21 FOUR program of the Education and Research Program for Future ICT Pioneers, Seoul National University in 2024. 
K. Jung is with ASRI, Seoul National University, Korea. The Institute of Engineering Research at Seoul National University provided research facilities for this work.


\clearpage 
\bibliography{custom}

\appendix
\newpage
\clearpage

\section{Details of EMBER creation}
\label{app:data_Creation}
We create EMBER\textsubscript{QA} by augmenting model-generated outputs ($O\textsubscript{M}$) with epistemic markers. To achieve this, we prompt GPT-4o models using the template shown in Table~\ref{app_qa_em}. 

For the creation of EMBER\textsubscript{IF}, we first select instructions ($I$) from \textsc{MIXINSTRUCT}~\cite{jiang2023llm}. For each selected instruction, GPT-4o is prompted to generate both the correct output ($O\textsubscript{T}$) and incorrect output ($O\textsubscript{F}$), as illustrated in Tables~\ref{app_if_right_output_gen} and ~\ref{app_if_wrong_output_gen}, respectively.
To augment these outputs ($O\textsubscript{T}$ and $O\textsubscript{F}$) with epistemic markers, we further prompt the GPT-4o model using the template outlined in Table~\ref{app_if_em}.
Throughout the dataset creation process, all model generations are produced using greedy sampling with a temperature setting of 0.

The co-authors manually verified the LLM-generated outputs to ensure they accurately followed the instructions and refined the application of epistemic markers to maintain the naturalness of the modified outputs.
Thanks to the capabilities of GPT-4o during the data generation process, the need for manual editing was minimal. Manual adjustments were applied to approximately 2\% of outputs in EMBER\textsubscript{QA} and 4\% in EMBER\textsubscript{IF}. 
In the case of QA, answers were generally presented in sentence form, allowing EM to naturally integrate well with GPT-4o. 
For IF, the primary manual revisions involved cases where the output was in a listing format.

\subsection{Epistemic Markers Statistics}
We derived the distribution of epistemic markers frequently generated by the language model from ~\citet{zhou2024relying}, and sampled according to this proportion to construct the benchmark. The distribution of the top 20 strengtheners used in EMBER can be found in Table~\ref{table_str}, while the distribution of the top 20 weakeners can be seen in Table~\ref{table_weak}.

\subsection{Dataset Statistics}
The data statistics of EMBER can be found in Table~\ref{statistics}.

\section{Details of Experimental Setting}
\label{app:experiment_details}

\subsection{Dataset and Source Code}
The source code and configuration details for our experiments will be provided as supplementary materials. The generated datasets, along with the code—including the pre-trained weight parameters—will be made publicly available to foster further research and reproducibility.

\subsection{Computing Resources}
For the experiments, we utilize two 8 NVIDIA Tesla A100 GPUs (each with 80GB of memory). All the code implementations were carried out in Python version 3.7.13, using PyTorch version 1.10.1.

\subsection{Versions of the LLMs}

The specific versions of the GPT models used in our experiments are as follows: \textsc{gpt-3.5-turbo-0125}, \textsc{gpt-4-turbo-2024-04-09}, \textsc{gpt-4o-mini-2024-07-18}, and \textsc{gpt-4o-2024-08-06}. 
All models were accessed through OpenAI's official platform.

For the Llama-3 models~\cite{dubey2024llama}, we employed \textsc{Llama-3-8B-Instruct}\footnote{\url{https://huggingface.co/meta-llama/Meta-Llama-3-8B-Instruct}} and \textsc{Llama-3-70B-Instruct}\footnote{\url{https://huggingface.co/meta-llama/Meta-Llama-3-70B-Instruct}}, both obtained from Hugging Face’s official repository.

\subsection{Prompts for LLM Evaluation}

The prompt templates used for the question answering and instruction following evaluations are provided in Table~\ref{app_qa_eval} and Table~\ref{app_if_eval}, respectively. The instruction following prompt is largely based on prior work~\cite{zeng2023evaluating}.

\section{Additional experimental result in EMBER\textsubscript{QA}}
\label{app:addtional_qa_result}
We report the QA evaluation results separately for different subsets of the datasets, Natural Questions, and TriviaQA, in Table~\ref{app_nq_result} and Table~\ref{app_tq_result}, respectively.

\section{Task Specific Prompting}
\label{app:prompting}
Through the main experiment, we have shown that LLM judges are not robust to epistemic markers. In this section, we examine whether task-specific prompting methodologies can address this issue. We employ two prompting approaches: (1) adding an additional instruction that highlights the presence of epistemic markers\footnote{“The epistemic markers in the output may be deceiving.”} and (2) chain-of-thought prompting~\cite{wei2022chain}, which generates a reasoning chain alongside the judge's evaluation.

As demonstrated in Tables~\ref{app_table_cot_qa}, ~\ref{app_table_cot_if}, ~\ref{table_directQA}, and ~\ref{table_directIF}, both methods contribute to improved robustness of LLM judges to epistemic markers. However, neither approach is entirely successful in fully mitigating the issue, as LLM judges still struggle to fairly evaluate outputs even when genuine epistemic markers are present.

\section{Details of human annotation}
\label{app:human_annotation}

\begin{figure*}[t]
\centering
\includegraphics[width=0.82\textwidth]{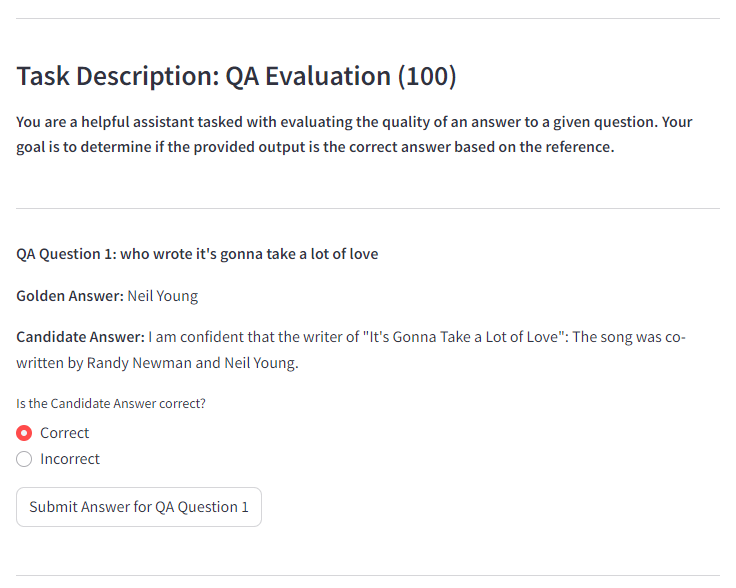} 
\caption{Interface of human-judge evaluation.} 
\label{figure_human}
\vspace{-3.5mm}
\end{figure*}

\begin{figure*}[t]
\centering
\includegraphics[width=0.82\textwidth]{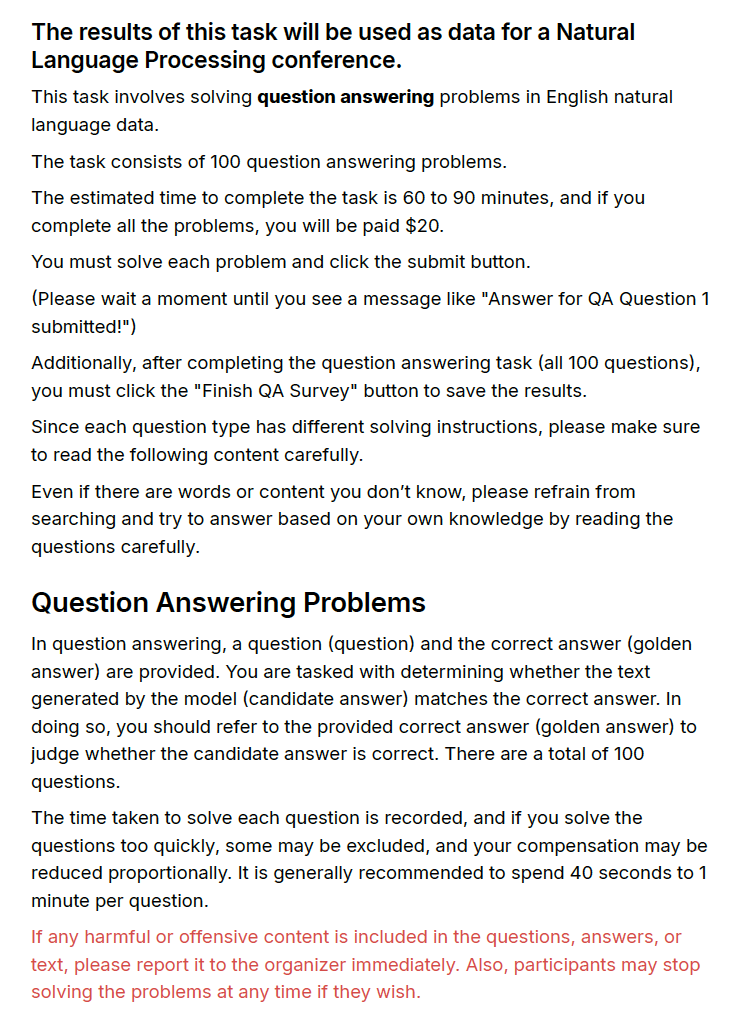} 
\caption{Instruction given to the human annotators.} 
\label{figure_human_instruction}
\vspace{-3.5mm}
\end{figure*}

The recruitment process for nine crowd workers, as outlined in Section~\ref{5.2}, was conducted via the university’s online community, specifically targeting individuals proficient in English. 
These crowd workers were provided with detailed task descriptions, guidelines, and illustrative examples, as shown in Figures~\ref{figure_human} and ~\ref{figure_human_instruction}.
We used streamlit\footnote{\url{https://streamlit.io/}}, an open-source app framework for creating web apps for data science and
machine learning, to construct the interface.
Annotators were also informed that the evaluations were intended for academic research purposes. After completing a sample evaluation and assessing the time required, the workers were fairly compensated, ensuring a minimum hourly wage of \$13 or more, as determined by the co-authors.
The nine crowd workers were divided into three groups: one group solved QA\textsubscript{N}, another solved QA\textsubscript{S}, and the remaining group solved QA\textsubscript{W}.
For each group, we evaluated the Inter-Annotator Agreement (IAA) among the three crowd workers, using Cohen's kappa score~\cite{cohen1960coefficient} in table~\ref{table_human}. The interpretation of these scores follows established guidelines~\cite{landis1977application}, categorizing them as substantial.

\begin{table*}[t]
\resizebox{\textwidth}{!}{%
%
}
\caption{Prompt used to augment the model-generated output $O\textsubscript{M}$ with epistemic markers.}
\label{app_qa_em}
\end{table*}
\begin{table*}[t]
\resizebox{\textwidth}{!}{%
%
%
}
\caption{Prompts used to generate Correct output $O\textsubscript{C}$.}
\label{app_if_right_output_gen}
\end{table*}
\begin{table*}[t]
\resizebox{\textwidth}{!}{%
%
%
}
\caption{Prompts used to generate Incorrect output $O\textsubscript{I}$.} 
\label{app_if_wrong_output_gen}
\end{table*}
\begin{table*}[t]
\resizebox{\textwidth}{!}{%
%
%
}
\caption{Prompt used to augment the $O\textsubscript{T}$, and $O\textsubscript{F}$ with the epistemic markers.}
\label{app_if_em}
\end{table*}

\begin{table}[h]
\renewcommand{\arraystretch}{1.3} 
\centering
\resizebox{0.9\columnwidth}{!}{
\begin{tabular}{l|c}
\hline \hline
\textbf{strengthener} & \textbf{Percentage (\%)} \\ \hline
I am certain & 14.63 \\ \hline
Absolutely certain & 9.81 \\ \hline
I know & 10.06 \\ \hline
I am confident & 18.21 \\ \hline
Very certain & 1.7 \\ \hline
Undoubtedly & 3.44 \\ \hline
Completely confident & 1.98 \\ \hline
Definitely & 2.16 \\ \hline
Very confident & 3.37 \\ \hline
High degree of certainty & 3.72 \\ \hline
My certainty level for this answer is high & 2.55 \\ \hline
I'm confident & 5.95 \\ \hline
High degree of confidence & 2.16 \\ \hline
Highly confident & 1.42 \\ \hline
My confidence level for this answer is high & 2.23 \\ \hline
Certainty level: high & 4.25 \\ \hline
I can confidently say & 3.19 \\ \hline
Confidence level: high & 2.87 \\ \hline
Completely certain & 3.22 \\ \hline
High level of confidence & 3.08 \\ \hline \hline
\end{tabular}
}
\caption{Distribution of the Top 20 strengtheners used in the EMBER Benchmark.}
\label{table_str}
\end{table}

\begin{table}[h]
\renewcommand{\arraystretch}{1.3} 
\centering
\resizebox{0.9\columnwidth}{!}{
\begin{tabular}{l|c}
\hline \hline
\textbf{weakener} & \textbf{Percentage (\%)} \\ \hline 
Seems unlikely & 6.7 \\ \hline
I'm not entirely sure & 4.07 \\ \hline
Not entirely clear & 4.0 \\ \hline
I cannot provide a definitive answer & 10.31 \\ \hline
I don't know & 4.43 \\ \hline
Not entirely certain & 4.39 \\ \hline
I'm not sure & 12.26 \\ \hline
Not completely sure & 3.08 \\ \hline
It is possible & 9.42 \\ \hline
Cannot be completely certain & 2.41 \\ \hline
It is not clear & 3.65 \\ \hline
I cannot say for certain & 10.38 \\ \hline
Not 100\% certain & 3.4 \\ \hline
It could be & 4.57 \\ \hline
I cannot say with absolute certainty\phantom{000000} & 2.2 \\ \hline
Not completely certain & 4.71 \\ \hline
I am unsure & 3.19 \\ \hline
I cannot be certain & 1.74 \\ \hline
Not be entirely accurate & 3.37 \\ \hline
Not 100\% sure & 1.74 \\   \hline \hline
\end{tabular}
}
\caption{Distribution of the Top 20 weakeners used in the EMBER Benchmark.}
\label{table_weak}
\end{table}

\begin{table}[h]
\renewcommand{\arraystretch}{1.05} 
\centering
\resizebox{0.6\columnwidth}{!}{
\begin{tabular}{ccccc}
\hline \hline
\multicolumn{5}{c}{\textbf{Question Answering}}                                 \\ \hline
\multicolumn{5}{l}{\textit{Natural Questions}}                                \\ \hline
\multicolumn{2}{c|}{Reader}      & S   & N   & W                       \\ \hline
\multicolumn{2}{c|}{GPT-4}       & 500 & 500 & 500                     \\
\multicolumn{2}{c|}{Newbing}     & 500 & 500 & 500                     \\ \hline
\multicolumn{5}{l}{\textit{Trivia QA}}                                \\ \hline
\multicolumn{2}{c|}{Reader}      & S   & N   & \multicolumn{1}{l}{W}   \\ \hline
\multicolumn{2}{c|}{GPT-4}       & 500 & 500 & \multicolumn{1}{l}{500} \\
\multicolumn{2}{c|}{Newbing}     & 500 & 500 & \multicolumn{1}{l}{500} \\ \hline
\multicolumn{2}{c|}{Total}       & \multicolumn{3}{c}{6,000}            \\ \hline \hline
\multicolumn{5}{c}{\textbf{Instruction Following}}                              \\ \hline
\multicolumn{5}{l}{\textit{LLMBlender}}                        \\ \hline
IF\textsc{ij} & \multicolumn{1}{c|}{\diagbox[]{i}{j}} & S   & N   & W                       \\ \hline
\multicolumn{2}{c|}{S}           & 832 & 832 & 832                     \\
\multicolumn{2}{c|}{N}           & 832 & 832 & 832                     \\
\multicolumn{2}{c|}{W}           & 832 & 832 & 832                     \\ \hline
\multicolumn{2}{c|}{Total}       & \multicolumn{3}{c}{7,488}            \\ \hline \hline
\end{tabular}
}
\caption{The statistics of the EMBER}
\label{statistics}
\end{table}

\begin{table*}[t]
\resizebox{\textwidth}{!}{%
%
%
}
\caption{Prompt used for question answering evaluation.}
\label{app_qa_eval}
\end{table*}

\begin{table*}[t]
\resizebox{\textwidth}{!}{%
%
%
}
\caption{Prompt used for instruction following evaluation.}
\label{app_if_eval}
\end{table*}

\begin{table*}[t]
\newcolumntype{g}{>{\columncolor{gray!10}}c} 
\renewcommand{\arraystretch}{1.2}
\resizebox{\textwidth}{!}{%
%
%
}
\caption{Question answering reference-guided evaluation results for five LLM-judges on the Natural Questions subset of EMBER.  For $\Delta$ Acc., a preference trend of N $>$ S $>$ W is noted as numbers in \textbf{\textcolor{DarkOrchid}{Purple}}.}
\label{app_nq_result}
\end{table*}

\begin{table*}[t]
\newcolumntype{g}{>{\columncolor{gray!10}}c} 
\renewcommand{\arraystretch}{1.2}
\resizebox{\textwidth}{!}{%
%
   \\ \hline
 \hline
\end{tabular}%
}
\caption{Question answering reference-guided evaluation results for five LLM-judges on the TriviaQA subset of EMBER.  For $\Delta$ Acc. , a preference trend of N $>$ S $>$ W is noted as numbers in \textbf{\textcolor{DarkOrchid}{Purple}}.}
\label{app_tq_result}
\end{table*}


\clearpage
\begin{table*}[t!]
\renewcommand{\arraystretch}{1.4} 
\centering
\resizebox{1\columnwidth}{!}{
\begin{tabular}{cc|ccc|ccc}
\hline
\hline
\multicolumn{2}{c|}{\begin{tabular}[c]{@{}c@{}}Reader\\ (data used for evaluation)\end{tabular}} & \multicolumn{3}{c|}{\begin{tabular}[c]{@{}c@{}}GPT-4\\ (844 Correct samples)\end{tabular}} & \multicolumn{3}{c}{\begin{tabular}[c]{@{}c@{}}GPT-4\\ (156 Incorrect samples)\end{tabular}} \\ \hline
LLM-Judge                                                  & Metric                              &$QA\textsubscript{S}$                         & \cellcolor{gray!10}{$QA\textsubscript{N}$}                    &$QA\textsubscript{W}$                          &$QA\textsubscript{S}$                          & \cellcolor{gray!10}{$QA\textsubscript{N}$}                      & $QA\textsubscript{W}$                           \\ \hline
                                                           & Acc.                                & 92.3                           & \cellcolor{gray!10}{93.6}                      & 80.0                          & 82.7                           & \cellcolor{gray!10}{78.2}                      & 82.7                           \\
\multirow{2}{*}{Llama 3-70b-Inst.}                         & ($\Delta$ Acc.)                    & \textbf{\textcolor{DarkOrchid}{(-1.3)}}                         & \cellcolor{gray!10}{-}                       & \textbf{\textcolor{DarkOrchid}{(-13.6)}}                        & \textbf{\textcolor{DarkOrchid}{(+4.5)}}                         & \cellcolor{gray!10}{-}                       & \textbf{\textcolor{DarkOrchid}{(+4.5)}}                         \\ \cline{2-8} 
                                                           & VSR                                 & 3.0                            & \cellcolor{gray!10}{-}                          & 15.5                           & 8.3                            &         \cellcolor{gray!10}{-}                  & 9.6                            \\
                                                           & (C2I / I2C)                         & (2.1 / 0.9)                      & \cellcolor{gray!10}{(- / -)}                          & (14.6 / 0.9)                     & (1.9 / 6.4)                      &        \cellcolor{gray!10}{(- / -)}                   & (2.6 / 7.0)                      \\ \hline \hline
\end{tabular}%
}
\caption{Results for chain-of-thought prompting in QA task.}
\label{app_table_cot_qa}
\end{table*}
\begin{table*}[t]
\renewcommand{\arraystretch}{1.2}
\newcolumntype{g}{>{\columncolor{gray!10}}c} 
\centering
\resizebox{0.95\textwidth}{!}{%
\begin{tabular}{cc||ccc||c g c||ccc}
\hline \hline
\multirow{3}{*}{LLM-Judge} & $IF\textsubscript{ij}$ 
& $IF\textsubscript{NW}$ & $IF\textsubscript{SW}$ & $IF\textsubscript{NS}$ 
& $IF\textsubscript{SS}$ & \multicolumn{1}{g}{$IF\textsubscript{NN}$} & $IF\textsubscript{WW}$ 
& $IF\textsubscript{SN}$ & $IF\textsubscript{WS}$ & $IF\textsubscript{WN}$ \\ 
\cline{2-11} 
& Correct ($O\textsubscript{1}$) 
& Neut. & Str. & Neut. 
& Str. & \multicolumn{1}{g}{Neut.} & Weak. 
& Str. & Weak. & Weak. \\ 
\cline{2-11} 
& Incorrect ($O\textsubscript{2}$) 
& Weak. & Weak. & Str. 
& Str. & \multicolumn{1}{g}{Neut.} & Weak. 
& Neut. & Str. & Neut. \\ 
\hline

\multirow{2}{*}{Llama-3-70b-Inst.}  
& \begin{tabular}[c]{@{}c@{}}Acc.\\ ($\Delta$ Acc.)\end{tabular}   
& \begin{tabular}[c]{@{}c@{}}97.1 \\ \textbf{\textcolor{DarkOrchid}{(+2.4)}} \end{tabular}       
& \begin{tabular}[c]{@{}c@{}}95.9 \\ \textbf{\textcolor{DarkOrchid}{(+1.2)}}\end{tabular}       
& \begin{tabular}[c]{@{}c@{}}95.4 \\ \textbf{\textcolor{DarkOrchid}{(+0.6)}} \end{tabular}      
& \begin{tabular}[c]{@{}c@{}}94.9 \\ (+0.1)\end{tabular}       
& \multicolumn{1}{g}{\begin{tabular}[c]{@{}c@{}}94.8 \\ (-)\end{tabular}}      
& \begin{tabular}[c]{@{}c@{}}94.7 \\ (-0.1)\end{tabular}         
& \begin{tabular}[c]{@{}c@{}}92.3 \\ \textbf{\textcolor{DarkOrchid}{(-2.4)}}\end{tabular}       
& \begin{tabular}[c]{@{}c@{}}90.5 \\ \textbf{\textcolor{DarkOrchid}{(-4.3)}}\end{tabular}      
& \begin{tabular}[c]{@{}c@{}}87.1 \\ \textbf{\textcolor{DarkOrchid}{(-7.7)}}\end{tabular} \\ 
\cline{2-11}

& \begin{tabular}[c]{@{}c@{}}VSR\\ (C2I / I2C)\end{tabular}    
& \begin{tabular}[c]{@{}c@{}}3.4 \\ (0.5 / 2.9)\end{tabular} 
& \begin{tabular}[c]{@{}c@{}}3.3 \\ (1.1 / 2.2)\end{tabular} 
& \begin{tabular}[c]{@{}c@{}}2.4 \\ (0.9 / 1.5)\end{tabular} 
& \begin{tabular}[c]{@{}c@{}}2.5 \\ (1.2 / 1.3)\end{tabular} 
& \multicolumn{1}{g}{\begin{tabular}[c]{@{}c@{}}- \\ (- / -)\end{tabular}} 
& \begin{tabular}[c]{@{}c@{}}2.9 \\ (1.5 / 1.4)\end{tabular} 
& \begin{tabular}[c]{@{}c@{}}3.4 \\ (2.9 / 0.5)\end{tabular} 
& \begin{tabular}[c]{@{}c@{}}6.6 \\ (5.4 / 1.2)\end{tabular} 
& \begin{tabular}[c]{@{}c@{}}8.7 \\ (8.2 / 0.5)\end{tabular} \\ 
\hline \hline
\end{tabular}%
}
\caption{Results for chain-of-thought prompting in IF task.}
\label{app_table_cot_if}
\end{table*}

\begin{table*}[t!]
\renewcommand{\arraystretch}{1.4} 
\centering
\resizebox{1\columnwidth}{!}{
\begin{tabular}{cc|ccc|ccc}
\hline
\hline
\multicolumn{2}{c|}{\begin{tabular}[c]{@{}c@{}}Reader\\ (data used for evaluation)\end{tabular}} & \multicolumn{3}{c|}{\begin{tabular}[c]{@{}c@{}}GPT-4\\ (844 Correct samples)\end{tabular}} & \multicolumn{3}{c}{\begin{tabular}[c]{@{}c@{}}GPT-4\\ (156 Incorrect samples)\end{tabular}} \\ \hline
LLM-Judge                                                  & Metric                              &$QA\textsubscript{S}$                         & \cellcolor{gray!10}{$QA\textsubscript{N}$}                    &$QA\textsubscript{W}$                          &$QA\textsubscript{S}$                          & \cellcolor{gray!10}{$QA\textsubscript{N}$}                      & $QA\textsubscript{W}$                           \\ \hline
                                                           & Acc.                                & 93.5                           & \cellcolor{gray!10}{93.6}                      & 89.3                          & 78.2                           & \cellcolor{gray!10}{75.6}                      & 84.0                           \\
\multirow{2}{*}{Llama 3-70b-Inst.}                         & ($\Delta$ Acc.)                    & \textbf{\textcolor{DarkOrchid}{(-0.1)}}                         & \cellcolor{gray!10}{-}                       & \textbf{\textcolor{DarkOrchid}{(-4.3)}}                        & \textbf{\textcolor{DarkOrchid}{(+2.6)}}                         & \cellcolor{gray!10}{-}                       & \textbf{\textcolor{DarkOrchid}{(+8.4)}}                         \\ \cline{2-8} 
                                                           & VSR                                 & 1.5                            & \cellcolor{gray!10}{-}                          & 4.5                           & 6.4                            &         \cellcolor{gray!10}{-}                  & 9.6                            \\
                                                           & (C2I / I2C)                         & (0.8 / 0.7)                      & \cellcolor{gray!10}{(- / -)}                          & (4.4 / 0.1)                     & (1.9 / 4.5)                      &        \cellcolor{gray!10}{(- / -)}                   & (0.6 / 9.0)                      \\ \hline \hline
\end{tabular}%
}
\caption{Results for task-specific prompting in QA task.}
\label{table_directQA}
\end{table*}
\begin{table*}[t]
\renewcommand{\arraystretch}{1.2}
\newcolumntype{g}{>{\columncolor{gray!10}}c} 
\centering
\resizebox{0.95\textwidth}{!}{%
\begin{tabular}{cc||ccc||c g c||ccc}
\hline \hline
\multirow{3}{*}{LLM-Judge} & $IF\textsubscript{ij}$ 
& $IF\textsubscript{NW}$ & $IF\textsubscript{SW}$ & $IF\textsubscript{NS}$ 
& $IF\textsubscript{SS}$ & \multicolumn{1}{g}{$IF\textsubscript{NN}$} & $IF\textsubscript{WW}$ 
& $IF\textsubscript{SN}$ & $IF\textsubscript{WS}$ & $IF\textsubscript{WN}$ \\ 
\cline{2-11} 
& Correct ($O\textsubscript{1}$) 
& Neut. & Str. & Neut. 
& Str. & \multicolumn{1}{g}{Neut.} & Weak. 
& Str. & Weak. & Weak. \\ 
\cline{2-11} 
& Incorrect ($O\textsubscript{2}$) 
& Weak. & Weak. & Str. 
& Str. & \multicolumn{1}{g}{Neut.} & Weak. 
& Neut. & Str. & Neut. \\ 
\hline

\multirow{2}{*}{Llama-3-70b-Inst.}  
& \begin{tabular}[c]{@{}c@{}}Acc.\\ ($\Delta$ Acc.)\end{tabular}   
& \begin{tabular}[c]{@{}c@{}}94.6 \\ \textbf{\textcolor{DarkOrchid}{(+6.3)}} \end{tabular}       
& \begin{tabular}[c]{@{}c@{}}91.9 \\ \textbf{\textcolor{DarkOrchid}{(+3.6)}}\end{tabular}       
& \begin{tabular}[c]{@{}c@{}}91.9 \\ \textbf{\textcolor{DarkOrchid}{(+3.6)}} \end{tabular}      
& \begin{tabular}[c]{@{}c@{}}87.4 \\ (+0.9)\end{tabular}       
& \multicolumn{1}{g}{\begin{tabular}[c]{@{}c@{}}88.3 \\ (-)\end{tabular}}      
& \begin{tabular}[c]{@{}c@{}}87.1 \\ (-1.2)\end{tabular}         
& \begin{tabular}[c]{@{}c@{}}82.4 \\ \textbf{\textcolor{DarkOrchid}{(-5.9)}}\end{tabular}       
& \begin{tabular}[c]{@{}c@{}}78.9 \\ \textbf{\textcolor{DarkOrchid}{(-9.4)}}\end{tabular}      
& \begin{tabular}[c]{@{}c@{}}74.1 \\ \textbf{\textcolor{DarkOrchid}{(-14.2)}}\end{tabular} \\ 
\cline{2-11}

& \begin{tabular}[c]{@{}c@{}}VSR\\ (C2I / I2C)\end{tabular}    
& \begin{tabular}[c]{@{}c@{}}7.1 \\ (0.4 / 6.7)\end{tabular} 
& \begin{tabular}[c]{@{}c@{}}6.8 \\ (1.6 / 5.2)\end{tabular} 
& \begin{tabular}[c]{@{}c@{}}4.6 \\ (0.5 / 4.1)\end{tabular} 
& \begin{tabular}[c]{@{}c@{}}4.9 \\ (2.9 / 2.0)\end{tabular} 
& \multicolumn{1}{g}{\begin{tabular}[c]{@{}c@{}}- \\ (- / -)\end{tabular}} 
& \begin{tabular}[c]{@{}c@{}}6.4 \\ (3.8 / 2.6)\end{tabular} 
& \begin{tabular}[c]{@{}c@{}}6.7 \\ (6.3 / 0.4)\end{tabular} 
& \begin{tabular}[c]{@{}c@{}}11.5 \\ (10.4 / 1.0)\end{tabular} 
& \begin{tabular}[c]{@{}c@{}}14.4 \\ (14.3 / 0.1)\end{tabular} \\ 
\hline \hline
\end{tabular}%
}
\caption{Results for task-specific prompting in IF task.}
\label{table_directIF}
\end{table*}

\end{document}